\documentclass{article}

\PassOptionsToPackage{numbers,sort&compress}{natbib}
\usepackage[final]{neurips_2022}
\input{al_header.tex}

\usepackage[utf8]{inputenc} %
\usepackage[T1]{fontenc}    %
\usepackage{url}            %
\usepackage{booktabs}       %
\usepackage{amsfonts}       %
\usepackage{nicefrac}       %
\usepackage{microtype}      %
\usepackage{xcolor}         %
\usepackage{wrapfig}
\usepackage{algorithm}
\usepackage{algorithmicx}
\usepackage[noend]{algpseudocode}
\usepackage{pifont}
\newcommand{\cmark}{\text{\ding{51}}}
\newcommand{\xmark}{\text{\ding{55}}}
\usepackage{DejaVuSans}
\usepackage{bbm}
\usepackage{stmaryrd}

\definecolor{ours}{HTML}{8fb030}
\definecolor{oursdarker}{HTML}{728c26}
\definecolor{baseline1}{HTML}{5c7fb3}
\definecolor{baseline2}{HTML}{e19c23}
\definecolor{baseline3}{HTML}{eb6133}
\definecolor{baseline4}{HTML}{8778b3}
\definecolor{baseline1darker}{HTML}{4d71a6}
\definecolor{cyan}{HTML}{00b2b2}

\usepackage[colorlinks=true, linkcolor=baseline1darker, 
filecolor=baseline1darker,
urlcolor=baseline1darker, citecolor=baseline1darker]{hyperref}       %

\usepackage{caption}
\usepackage{subcaption}

\newcommand{\FPP}{\textsc{Construction}\xspace}
\newcommand{\Playground}{\textsc{Playground}\xspace}
\newcommand{\ourMethod}{CEE-US\xspace}
\newcommand{\mlpicem}{\mbox{MLP\,+\,iCEM}\xspace}
\newcommand{\Robodesk}{\textsc{RoboDesk}\xspace}

\title{Curious Exploration via Structured World Models Yields Zero-Shot Object Manipulation}

\author{%
  Cansu Sancaktar \\
  \And
  Sebastian Blaes \\
  \And
  Georg Martius \\
  \AND
  \normalfont{Max Planck Institute for Intelligent Systems} \\
  Tübingen, Germany\\
  \texttt{\{cansu.sancaktar, sebastian.blaes, georg.martius\}@tue.mpg.de} \\
}

\begin{document}
\maketitle
\begin{abstract}
  It has been a long-standing dream to design artificial agents that explore their environment efficiently via intrinsic motivation, similar to how children perform curious free play. Despite recent advances in intrinsically motivated reinforcement learning (RL), sample-efficient exploration in object manipulation scenarios remains a significant challenge as most of the relevant information lies in the sparse agent-object and object-object interactions. In this paper, we propose to use structured world models to incorporate relational inductive biases in the control loop to achieve sample-efficient and interaction-rich exploration in compositional multi-object environments. By planning for future novelty inside  structured world models, our method generates free-play behavior that starts to interact with objects early on and develops more complex behavior over time. Instead of using models only to compute intrinsic rewards, as commonly done, our method showcases that the self-reinforcing cycle between good models and good exploration also opens up another avenue: zero-shot generalization to downstream tasks via model-based planning. 
  After the entirely intrinsic task-agnostic exploration phase, our method solves challenging downstream tasks such as stacking, flipping, pick \& place, and throwing and generalizes to unseen numbers and arrangements of objects without any additional training.\footnote{Code and videos are available at \url{https://martius-lab.github.io/cee-us}.}
\end{abstract}

\section{Introduction}

Curious free-play has been identified as a driving force in child development, allowing children to efficiently explore their environment and build an understanding of the world \cite{loewenstein1994psychology}. 
Such intrinsically motivated exploration schemes are especially attractive in open-ended learning scenarios to guide an agent even without extrinsic tasks and corresponding rewards. Similar to how children learn, we want Reinforcement Learning (RL) agents to learn through play and then be able to solve new tasks quickly. 
We address both challenges in this work.

Minimization of novelty (or surprise) is a prominent formulation of curiosity, with several psychological studies showcasing the role of novelty in children's curious exploration \cite{legare_inconsistency_2010, bonawitz_children_2012, stahl2015observing}. Novelty as an intrinsic reward signal has also been adopted in RL, where agents try to resolve a cognitive disequilibrium \cite{schmidhuber1991possibility, pathak2017curiosity}. %
However, applying entirely intrinsic task-agnostic exploration to object manipulation scenarios in a sample-efficient manner is an ongoing challenge as the relevant information lies in the sparse agent-object and object-object interactions. These compositional multi-object manipulation environments highlight one of the significant weaknesses of the current novelty-based intrinsic motivation methods: a novel stimulus alone does not necessarily mean that it contains useful or generalizable information for an individual \cite{dubey2017rational}. 
Thus, curious exploration needs to attend to a subset of possibilities, as supported by studies in psychology \cite{kidd2015psychology,poli2020infants}.
Analyzing children during free play accumulated evidence that infants have innate biases and heuristics to help guide their attention toward relevant 
and informative 
features of the environment \cite{haith1980rules, murphy1985role}.
Our goal is to improve curious exploration in RL by incorporating appropriate inductive biases. We hypothesize that viewing the world as a collection of entities and their interactions is such a ``useful'' inductive bias, which was also put forward in \citet{tsividis2021human}. %
However, instead of utilizing explicit theory-based modeling, we exploit a relational inductive bias by using Graph Neural Networks (GNN) as our choice of model \cite{battaglia2018relational} and learn everything from interactions.
We postulate that curious exploration via epistemic uncertainty of such a model leads to the collection of valuable data as it inherits this structure.

This paper shows how online planning methods can efficiently exploit learned models, both for exploration and zero-shot generalization to tasks.
Recent advances in general-purpose model predictive control methods have reduced their complexity and show strong performance when good forward models are available \cite{pinneri2020:iCEM, VlastelicaBlaesEtal2021:riskaverse}. 
We build on this work to create a self-reinforcing cycle between learning good models and good exploration.
Thanks to their ability to plan for a new task without further training,
 models efficiently track the naturally changing exploration targets and perform downstream tasks in a zero-shot generalization manner.

We propose \ourMethod: \textbf{C}urious \textbf{E}xploration using \textbf{E}pistemic \textbf{U}ncertainty via \textbf{S}tructured Models that achieves sample-efficient and interaction-rich exploration in multi-object manipulation environments. 
To our knowledge, we are the first to 
use GNN ensemble disagreement for computing intrinsic motivation signals in RL. A further major contribution of our method is to combine GNN-based epistemic uncertainty with planning methods in a task-agnostic setting and demonstrate zero-shot generalization to challenging object manipulation tasks.
We provide a brief comparison to related work in \tab{tab:overview} and provide more details in \sec{sec:related}.

\begin{figure}
  \centering
  \includegraphics[width=\textwidth]{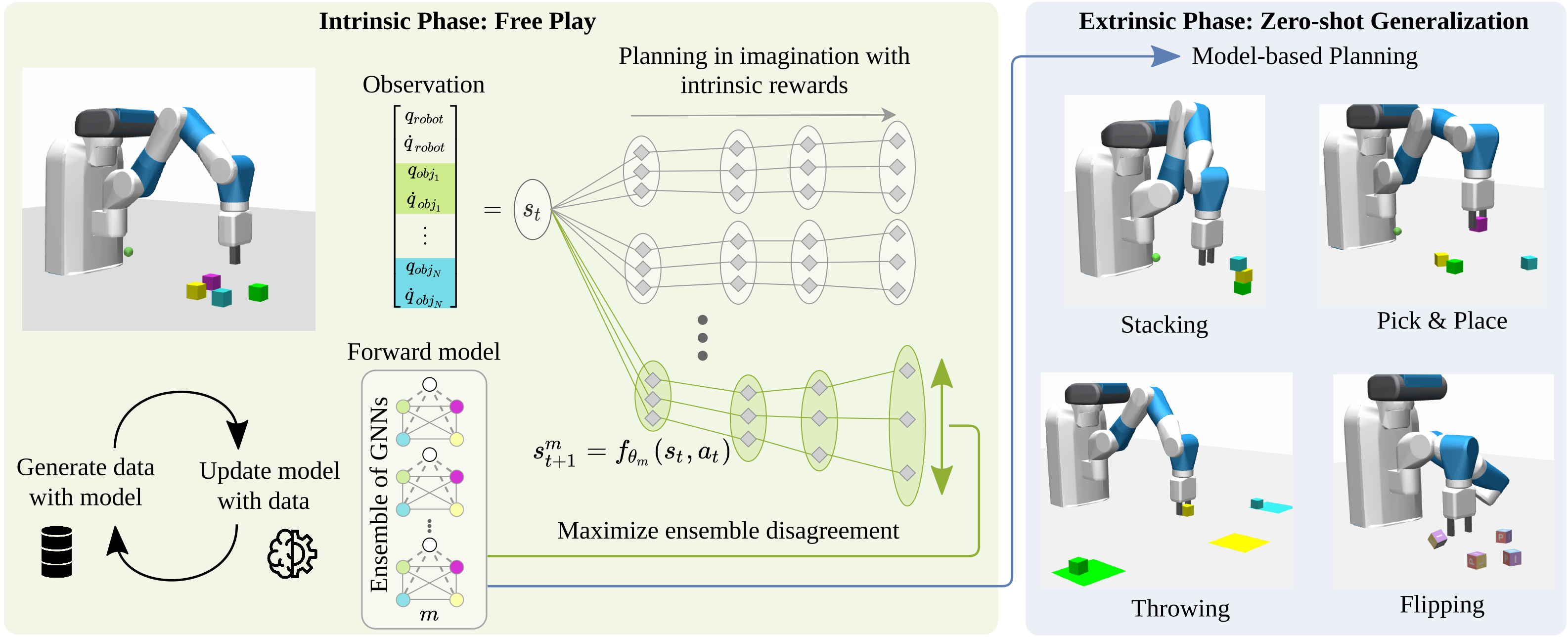}
  \caption{{\bf Intrinsic motivation for epistemic uncertainty reduction with structured world models yields capable models for zero-shot task planning.} 
  In an intrinsic phase, the system plans for maximum epistemic uncertainty with an ensemble of GNNs and uses the actively collected data to update the model. This iterative procedure ultimately reduces the overall epistemic uncertainty of the system, \ie maximizing information gain. The learned models can be used to solve complex manipulation tasks using online planning.}
\label{fig:overview} 
\end{figure}

\begin{table}[htb]
    \centering
    \caption{{\bf Comparison of \ourMethod with baselines} in terms of differences in paradigms.}
    \label{tab:overview}
    \resizebox{.9\linewidth}{!}{ %
    \begin{tabular}{@{}l|ccccc@{}}
    \toprule
    Method & \shortstack{Observation type} & \shortstack{Planner \\ or policy} &\shortstack{Learns \\ dynamics} & \shortstack{ Zero-shot task\\ generalization}& \shortstack{Combinatorial\\ generalization}\\
    \midrule
    \ourMethod & Proprioceptive & Planner &\cmark & \cmark & \cmark \\
    \mlpicem & Proprioceptive & Planner &\cmark & \cmark & \xmark \\
    Disagreement \cite{pathak2019self} & Propr. + Image & Policy &\cmark & \xmark & \xmark \\
    RND \cite{burda2018exploration}& Propr. + Image & Policy &\xmark & \xmark & \xmark \\
    ICM \cite{pathak2017curiosity}& Propr. + Image & Policy &\cmark & \xmark & \xmark \\
    Plan2Explore \cite{sekar2020planning}& Image & Policy & \cmark & \cmark (offline RL) & \xmark \\
   \bottomrule
    \end{tabular}
    }
\end{table}

\section{Method} \label{sec:method}
We focus on intrinsically motivated learning to prepare for future tasks.
Thus, we have a task-agnostic RL setting without extrinsic rewards or any other information regarding future tasks/goals during the initial play phase. 
Our approach trains a structured world model to capture the forward dynamics of the environment from active exploration data. In particular, we are using a Graph Neural Network (GNN) where the nodes correspond to objects.
We rely on our model's epistemic uncertainty to direct exploration during free play. Our focus is on compositional multi-object environments, where an actuated agent can independently manipulate different objects. The terms (actuated) agent and robot are used interchangeably. 

\subsection{Preliminaries} \label{sec:preliminaries}
In this work, we consider a fully observable Markov Decision Process (MDP) setting,
The MDP is given by \( \mathcal{M} = (\mathcal{S}, \mathcal{A}, P_{ss^{\prime}}^{a}, R_{ss^{\prime}}^{a}) \), with the continuous state-space \( \mathcal{S} \in \mathbb{R}^{n_{s}} \), the continuous action-space \( \mathcal{A} \in \mathbb{R}^{n_{a}} \), the transition kernel \( P_{ss^{\prime}}^{a} \), and the reward function \( R_{ss^{\prime}}^{a} \). 
Furthermore, we consider an object-oriented state representation, \ie the state-space factorizes into the different entities in the environment $\mathcal{S} = \mathcal{S}_1 \times \mathcal{S}_2 \dots \times \mathcal{S}_N \times \mathcal{S}_{\mathrm{agent}}$, where $N$ denotes the number of objects. 

In the typical RL setting, the agent's goal is to learn a policy \( a \sim \pi(\cdot \mid s) \) that maximizes the future (discounted) cumulative reward \( G_t = \sum_{k=0}^{\infty} \gamma^k \cdot R_{t+k} \), with the discount factor \( 0 < \gamma \leq 1 \). 
Such a policy can be learned as a neural network with RL algorithms, or a planning method can be used to maximize the same quantity. 

\subsubsection{Planning and Model Predictive Control} \label{sec:mpc}
Planning methods use a model \(f\) of the transition kernel \( P_{ss^{\prime}}^{a} \) and a reward function \( R_{ss^{\prime}}^a \) to optimize a sequence of actions on the fly. 
The potential advantage of planning methods is that they can optimize for different reward functions without further adaptation.
For this to be successful, a good transition model $f$ needs to be learned, and the reward function needs to be known or discovered.
More formally, for a fixed planning horizon \(H\), the action sequence \( \mathbf{a}_t = (a_{t+h})_{h=0}^{H-1} \) 
is optimized to maximize
\begin{equation} 
    \mathbf{a}_t^{\star} = \argmax_{\mathbf{a}_t} \sum_{h=0}^{H-1} R(s_{t+h}, a_{t+h}, s_{t+h+1}), \label{eqn:action_planner} 
\end{equation}
where \(s_{t+h}\) are imagined states visited by rolling out the actions using \(f\), which is assumed to be deterministic. 
For this procedure to optimize the same infinite horizon reward with discounting, $\gamma$ and a value function $V(s_{t+H})$ could be added to \eqn{eqn:action_planner}, sacrificing the flexibility in exchanging $R$.

We use zero-order trajectory optimization to find actions according to \eqn{eqn:action_planner} and model predictive control (MPC) to convert the open-loop planning policy into a closed-loop policy (re-planning after every step in the environment). In particular, we use the improved Cross-Entropy Method (iCEM)~\cite{pinneri2020:iCEM}, which was recently proposed as a sample efficient trajectory optimizer.

\subsection{World Model with Graph Neural Networks} \label{sec:gnn}
As the forward dynamics $f: \mathcal{S} \times \mathcal{A} \rightarrow \mathcal{S}$ (world model) is an integral part of the planning procedure, it needs to be able to capture the true dynamics well. 
We employ an ensemble of message-passing GNNs \cite{battaglia2018relational}.
Each object corresponds to a node in the graph, and the node attributes $\{s_t^{n} \in \mathcal{S}_n \mid n=1,\ldots,N\}$ are given by an object's features such as position, orientation, and velocity at timestep $t$. The state representation of the actuated agent $s^{\text{agent}} \in \mathcal{S}_{\text{agent}}$ similarly contains position and velocity information about the robot.

The agent representation is differentiated from the object nodes as it has a direct cause-and-effect relationship with the actions. 
The agent's state and the generic action are represented as a global context $c$.
The GNN is fully connected, \ie there is an edge between all nodes. 
In GNNs, the node update function $g_{\text{node}}$ models the dynamics of individual entities, and the edge update function $g_{\text{edge}}$ captures their pairwise interactions. 

The node attribute computation for the objects is given by:
\begin{align}
    c &= [s_t^{\text{agent}}, a_t] \\
    e_t^{(i,j)} &= g_{\text{edge}}\big(\big[s_t^i, s_t^j, c\big]\big) \\
    \hat{s}_{t+1}^{i} &= g_{\text{node}}\big(\big[s_t^i, c, \operatorname{aggr}_{i \neq j}\big(e_t^{(i,j)}\big)\big]\big).
\end{align}
where $[\cdot,\ldots]$ denotes concatenation, $e_t^{(i,j)}$ is the edge attribute between two neighboring nodes $(i, j)$, and $c$ is the global context information. The permutation-invariant aggregation function is given by $\operatorname{aggr}$. 
The $g_{\text{node}}$ and $g_{\text{edge}}$ are both Multilayer Perceptrons (MLP).

The context update, \ie the transition of the agent's state, is computed using the global aggregation of all edges similar to the global context formalism in \citet{battaglia2018relational}:
\begin{equation}
     \hat{s}_{t+1}^{\text{agent}} = g_{\text{global}}\big(\big[c, 
     \operatorname{aggr}_{i,j}\big(e_t^{(i,j)}\big)\big]\big),
     \label{}
\end{equation}
where $\mathbf{g}_{\text{global}}$ denotes the global node MLP. 
By providing all information to the prediction of the agent, we 
ensure an accurate modeling of agent-object interactions and their influence on the agent, which is paramount in object manipulation environments.
Using this representation, we found that a single message passing step is sufficient, thus yielding fast inference times.

Moreover, to make the model focus on capturing the changes accurately,
 we let the GNN predict $\Delta\hat{s}_{t+1} = \hat{s}_{t+1} - s_t$, instead of the absolute next state, such that $f(s,a) = s + GNN(s,a)$ %

\subsection{Epistemic Uncertainty as Intrinsic Reward}
We train an ensemble of GNNs $\{({f_{\theta}}_m)_{m=1}^M\}$ (\fig{fig:overview}), where $M$ denotes the ensemble size. %
The epistemic uncertainty, \ie the uncertainty due to lack of data, can be approximated by the disagreement of the ensemble members' predictions measured by the trace of the covariance matrix \cite{VlastelicaBlaesEtal2021:riskaverse}:
\begin{equation}
R_\mathrm{I}(s_t,a_t,\cdot) = \mathrm{tr}\big(\mathrm{Cov}(\{\hat{s}^m_{t+1}={f_{\theta}}_m(s_t, a_t) \mid m=1,\ldots,M\} )\big). \label{eqn:epistemic}
\end{equation}

\paragraph{Connections to Information Gain}
If more data becomes available from a region where the ensemble members disagree, then we expect the disagreement in this region to shrink, given enough capacity of the predictors. By guiding the exploration to regions with large model disagreement or high epistemic uncertainty, we explicitly aim for maximizing information gain \cite{Pfaffelhuber1972:MissingInformation,Storck1995Reinforcement-Driven-Information, Cover2006Elements-of-Information}.
Interestingly, as we can \emph{predict} the uncertainty for unvisited states, our approach is more related to predicted information gain \cite{LittleSommer2013:PIG}.

\begin{figure}
  \centering
  \includegraphics[width=\textwidth]{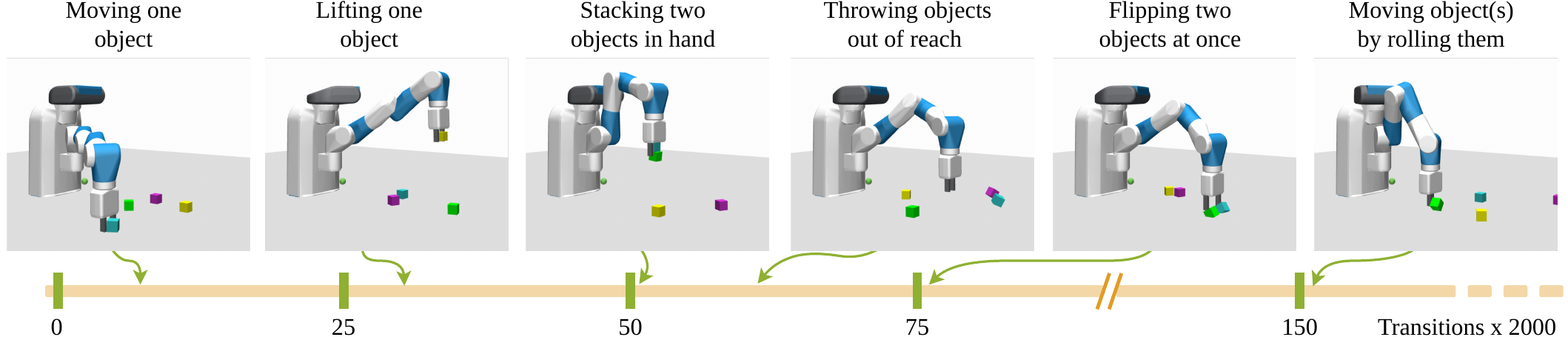}
  \caption{{\bf Emergent behaviors observed during free play} with \ourMethod in \FPP. Along the time axis, 2000 transitions correspond to one training iteration of \ourMethod, with 20 rollouts (episodes) of length 100. After only 30 iterations of free play, the robot starts lifting objects.}
  \label{fig:emergent:behavior}
\end{figure}

\subsection{The \ourMethod Algorithm} \label{sec:algorithm}
\ourMethod has two different phases: the intrinsic free-play phase (\alg{alg:intrinsic}), where learning occurs with active exploration, and the extrinsic phase (\alg{alg:extrinsic}), where we apply the learned model to solve downstream tasks zero-shot without any additional training.

\paragraph{Intrinsic Phase} In each iteration, the zero-order trajectory optimizer is used to plan for action sequences with high cumulative epistemic uncertainty in the imagined trajectories of the world model. The collected rollouts are added to the buffer, and the ensemble members of the world model are trained on the observed state transitions $(\mathbf{s}_{t}, \mathbf{a}_{t}, \mathbf{s}_{t+1})$ for a fixed number of epochs minimizing the loss function:
\begin{equation}
    \mathcal{L}_m = \norm{\Delta s_{t+1}-{f_\theta}_m(s_t, a_t)}_2^2
\end{equation}
for each ensemble member $m$ with independently sampled mini-batches, where $\Delta s_{t+1}=s_{t+1}-s_t$ denotes the actual change in the next state observation.
The total loss function is given by: $\mathcal{L}=\sum_{i=0}^M\mathcal{L}_m$.
Afterwards, the process of data collection and training is repeated. 
The pseudocode for the intrinsic phase is provided in \alg{alg:intrinsic}. Lines 3--9 correspond to one training iteration. %
\algrenewcommand\algorithmicindent{1em}
\algrenewcommand{\algorithmiccomment}[1]{\bgroup\hskip1em\textcolor{ourred}{\hfill$\rhd$~\textsl{#1}}\egroup}
\algrenewcommand{\alglinenumber}[1]{\color{gray}\scriptsize #1:}
\begin{minipage}[t]{.49\linewidth}
\captionsetup[algorithm]{font=small}
\begin{algorithm}[H]
  \scriptsize
  \caption{\bf \ourMethod: {\color{ourgreen}Free Play in Intrinsic Phase}} \label{alg:intrinsic}
  \begin{algorithmic}[1]
    \State \textbf{Input:} $\{({f_{\theta}}_m)_{m=1}^M\}$: Randomly initialized ensemble of GNNs with $M$ members, $D$: empty dataset, \texttt{Planner}: iCEM planner with horizon $H$
      \While{explore}\Comment{Explore with MPC and intrinsic reward}
        \For{$e = 1$ \textbf{to} $\texttt{num\_episodes}$}
            \For{$t = 1$ \textbf{to} $T$} \Comment{Plan to maximize model uncertainty}
                \State $a_t \gets \texttt{Planner}(s_t, \{({f_{\theta}}_m)_{m=1}^M\}, R_\mathrm{I})$ \Comment{\eqn{eqn:epistemic}}
                \State $s_{t+1} \gets \texttt{env.step}(s_t, a_t)$
            \EndFor
            \State $\mathcal{D} \leftarrow \mathcal{D} \cup \{(s_t, a_t, s_{t+1})_{t=1}^T\}$
        \EndFor
      \For{$l = 1$ \textbf{to} $L$}\Comment{Train models on dataset for $L$ epochs}
         \State $\theta_m \gets$ optimize $\theta_m$ using $\mathcal L_m$ on $\mathcal D$ for $m=1,\dots,M$%
      \EndFor
      \EndWhile
      \State \textbf{return} $\{({f_{\theta}}_m)_{m=1}^M\}, \mathcal{D}$
  \end{algorithmic}
\end{algorithm}
\end{minipage}
\hfill
\begin{minipage}[t]{.49\linewidth}
\captionsetup[algorithm]{font=small}
\begin{algorithm}[H]
    \scriptsize
  \caption{\bf \ourMethod: {\color{ourblue}Zero-shot Generalization in Extrinsic Phase}} \label{alg:extrinsic}
  \begin{algorithmic}[1]
    \State \textbf{Input:} $\{({f_{\theta}}_m)_{m=1}^M\}$: Ensemble of GNNs with $m$ members learned during intrinsic phase, \texttt{Planner}: iCEM planner with horizon $H$, $R_\text{task}$: Reward function of extrinsic task
      \While{\textbf{not} done} \Comment{Plan to maximize task reward in model}
        \State $a_t \gets \texttt{Planner}(s_t, \{({f_{\theta}}_m)_{m=1}^M\}, R_\text{task})$ \Comment{task reward}
        \State $s_{t+1} \gets \texttt{env.step}(s_t, a_t)$
      \EndWhile
  \end{algorithmic}
\end{algorithm}
\end{minipage}

\paragraph{Extrinsic Phase} The learned world model is used to perform planning for several downstream tasks, which we assume to be given in terms of reward functions $R_{\text{task}}(s,a,s^{\prime})$. The pseudocode is presented in \alg{alg:extrinsic}. Here, no further adaptation to the model is performed, although our method could be extended to incorporate a potential fine-tuning procedure.

\section{Experiments}
In our empirical evaluation, we analyze the performance of \ourMethod{} on two different object manipulation environments 
to answer the questions: 
How much does the structural inductive bias introduced by GNNs help model learning and control? 
Does the free-play phase create rich interaction data that helps downstream task performance? 
Can we solve challenging manipulation tasks in a zero-shot manner? 
The two environments we consider are: 

\paragraph{Playground} An actuated spherical agent can slide in the x-y directions and push four different object types (light cube, heavy cube, pyramid, and cylinder) along the x-y directions and also rotate them around the z-axis (\fig{fig:gnn_vs_mlp}). Each object has 3 Degrees of Freedom (DoF). Object types are uniquely identified by their color.

\paragraph{Fetch Pick \& Place Construction} This is an extension of the Fetch Pick \& Place environment~\cite{Plappert2018MultiGoalRL} to more cubes~\cite{li2020towards} (\fig{fig:emergent:behavior}). A 7-DoF robot arm is used to manipulate blocks (each with 6 DoF). The actions $a \in \mathbb{R}^4$ control the gripper movement in Cartesian coordinates and the gripper opening/closing. The robot state $\mathbf{s}_{\text{agent}} \in \mathbb{R}^{10}$ contains the end-effector position and velocity and the gripper-state (open/close) and gripper-velocity.
Each object's state $s^i \in \mathbb{R}^{12}$ is given by its position, orientation (in Euler angles), and linear and angular velocities. We replaced the table in front of the robot with a large plane so that objects cannot fall off during free play. However, they can still be pushed or thrown outside of the manipulability range of the robot.
Originally, each object state contained the object's position relative to the gripper. We remove this privileged information from the objects' state, as it already introduces a relational bias in the raw state representation.

\subsection{Structured vs.\ Unstructured World Models}

\begin{figure}
\center
\includegraphics[width=\textwidth]{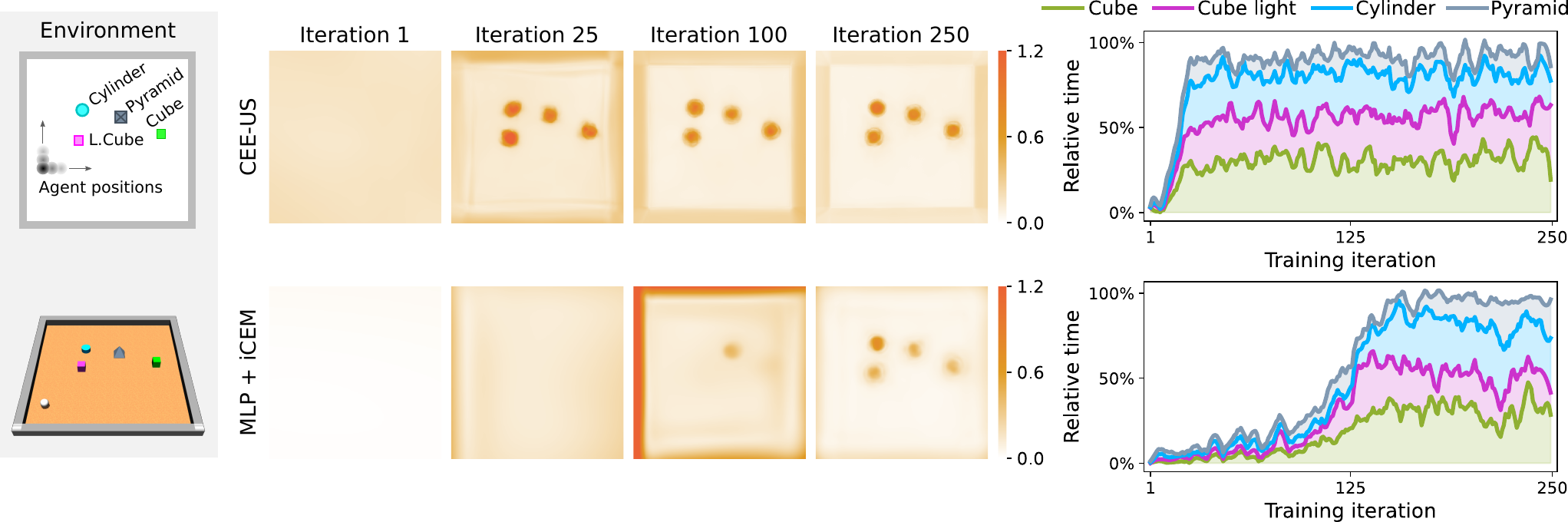}\vspace{-.5em}
\caption{\textbf{Structured world models lead to more object interactions.} 
In the playground environment, (left) we compare active exploration with an ensemble of GNNs (top row) vs.\ an ensemble of MLPs (bottom row) using epistemic uncertainty as an intrinsic reward ($M=5$). 
Middle: The heatmaps show the uncertainty of the ensemble members for hypothetical agent positions over a spatially discretized grid at different training iterations (see \supp{app:uncertainty_heatmap}).
Right: The relative time (the fraction of time steps) the agent spends moving the different objects in the playground is illustrated throughout free play.}
\label{fig:gnn_vs_mlp}
\end{figure}

We analyze how curious exploration based on minimizing epistemic uncertainty performs when using ensembles of GNNs and ensembles of MLPs (\fig{fig:gnn_vs_mlp}). MLPs, with their fully connected layers and monolithic input representations, do not offer isolation of information and incorporate no explicit relational inductive bias \cite{battaglia2018relational}, thus constituting a good baseline for our structured world models.
Results in \Playground{} show that the GNN ensemble leads much faster to interaction-rich data than its MLP counterpart. As visualized in \fig{fig:gnn_vs_mlp}, the uncertainty produced by the GNN ensemble is already localized around objects after 25 training iterations,
 leading to targeted agent-object and object-object interactions.
For the MLP, it takes more than 100 iterations to start generating useful uncertainty estimates and therefore object-agent interactions.
Also, note the pronounced uncertainty at the walls for the MLP. %
A more fine-grained analysis of the interaction times of \ourMethod compared to the baselines will be provided below. In \supp{app:gnn_pred}, we also show how the resulting multi-step dynamics predictions with structured world models are more accurate compared to the MLPs. 
This is a key component for the self-reinforcing cycle between good models and good exploration as a better model means: the agent can plan for more complex behavior earlier and learn from these experiences faster. 

\subsection{Interaction-Richness of Generated Exploration Data}
\label{interaction}
We perform an analysis of the data generated during intrinsic free-play. We compare the performance of our method to the unstructured \mlpicem variant (\ourMethod without GNNs) as well as other intrinsic motivation baselines: 

\textbf{Disagreement \cite{pathak2019self}}\quad
The one-step disagreement of an MLP ensemble is used as an intrinsic reward to train an exploration policy.

\textbf{Random Network Distillation \cite{burda2018exploration}}\quad  A predictor network, corresponding to the forward model, tries to match the output of a target network with random weights. The discrepancy between the two networks is used as an intrinsic reward (a type of state visitation count for continuous domains).%
    
\begin{figure}[b]
    \centering
    {\scriptsize
      \textcolor{ours}{\rule[2pt]{20pt}{1pt}} \ourMethod{} \quad \textcolor{baseline1}{\rule[2pt]{20pt}{1pt}} \mlpicem \quad \textcolor{baseline2}{\rule[2pt]{20pt}{1pt}} Disagreement \quad \textcolor{baseline3}{\rule[2pt]{20pt}{1pt}} RND
       \quad \textcolor{baseline4}{\rule[2pt]{20pt}{1pt}} ICM
    }\\
    \begin{subfigure}[b]{0.01\textwidth}
        \notsotiny{\rotatebox{90}{\hspace{.5cm}\textsf{relative time}}}
    \end{subfigure}      
    \begin{subfigure}[t]{.24\linewidth}
    \includegraphics[width=\linewidth]{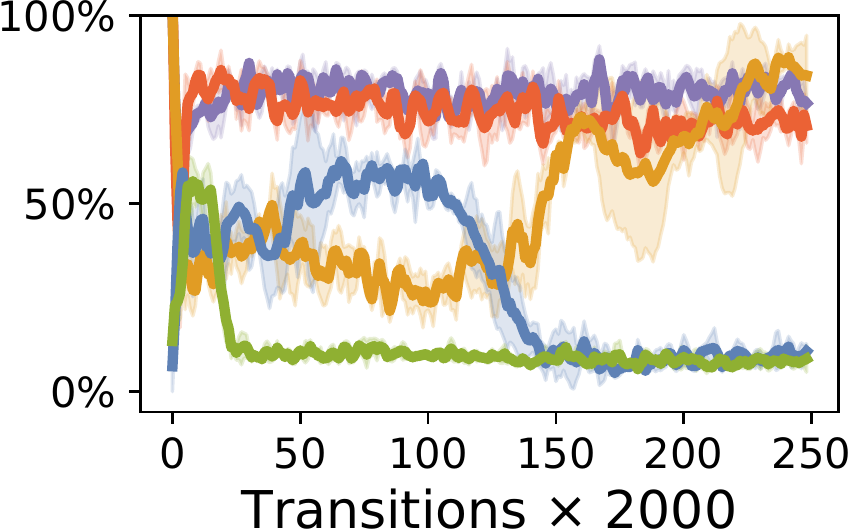}\vspace{-.3em}
    \caption{Agent in free space}
    \end{subfigure}
    \hfill
    \begin{subfigure}[t]{.24\linewidth}
    \includegraphics[width=\linewidth]{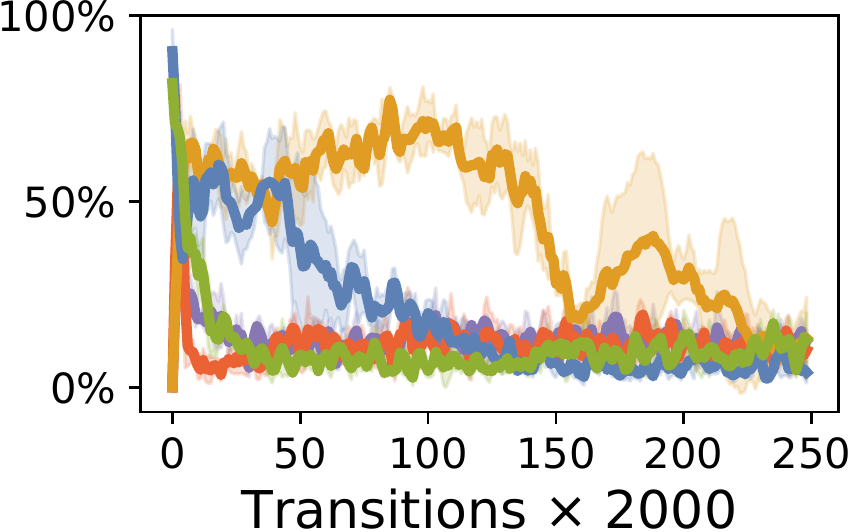}\vspace{-.3em}
    \caption{Agent at wall}
    \end{subfigure}
    \hfill
    \begin{subfigure}[t]{.24\linewidth}
    \includegraphics[width=\linewidth]{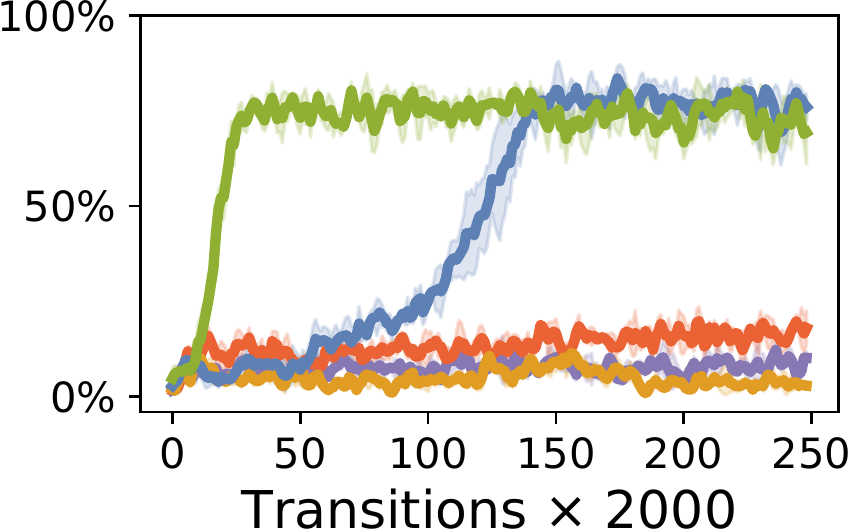}\vspace{-.3em}
    \caption{1 object moves}
    \end{subfigure}
    \hfill
    \begin{subfigure}[t]{.24\linewidth}
    \includegraphics[width=\linewidth]{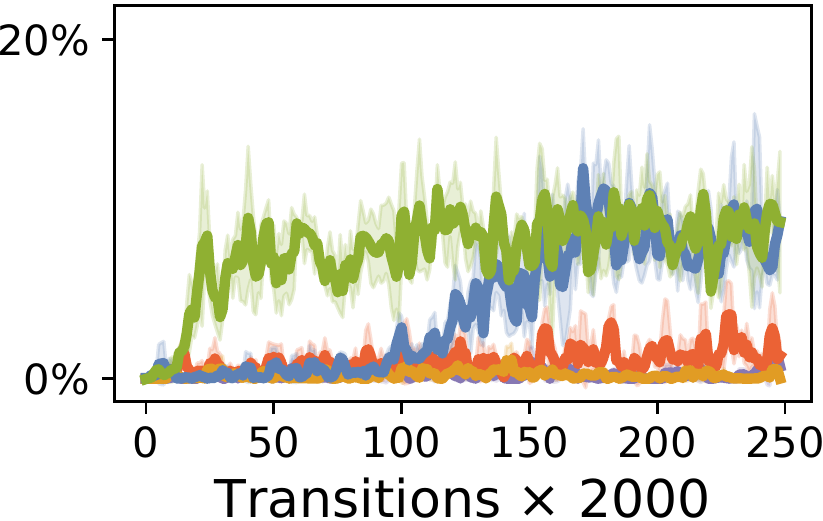}\vspace{-.3em}
    \caption{2 or more objects move}
    \end{subfigure}
    \caption{{\bf Interaction rate during intrinsic exploration in the \Playground environment.}
    The relative interaction times show that \ourMethod{} quickly starts interacting with the objects. \mlpicem is similar but one order of magnitude less data efficient. The baselines rarely engage in interesting interactions, mostly staying in free space or at the walls (three independent seeds).
    }
    \label{fig:interactions:playground}
\end{figure}

\textbf{ICM \cite{pathak2017curiosity}}\quad The intrinsic reward is defined as the error between an MLP forward model's next state prediction and the actual next state. As this method needs access to the true next state, the intrinsic reward can only be computed retrospectively.

The intrinsic reward is used to train an exploration policy in these methods. We use the implementation from \citet{laskin2021urlb} that uses DDPG \cite{Lillicrap2016DDPG}.
In \fig{fig:interactions:playground}, we present different metrics quantifying the amount and type of interactions that occur during the free play in the \Playground environment.
It is eminent that \ourMethod is only shortly interacting with the walls and then quickly interacts with one and two objects simultaneously. The next best method is the ablation of our method using MLP models needing about ten times more interactions, as explained in \fig{fig:gnn_vs_mlp}. 
The baselines are mainly moving in free space and interacting with the walls. RND starts interacting with the objects in at least 10\% of the times-steps.

\begin{figure}
    \centering
    {\scriptsize
      \textcolor{ours}{\rule[2pt]{20pt}{1pt}} \ourMethod{} \quad \textcolor{baseline1}{\rule[2pt]{20pt}{1pt}} \mlpicem \quad \textcolor{baseline2}{\rule[2pt]{20pt}{1pt}} Disagreement \quad \textcolor{baseline3}{\rule[2pt]{20pt}{1pt}} RND
       \quad \textcolor{baseline4}{\rule[2pt]{20pt}{1pt}} ICM
    }\\
    \begin{subfigure}[b]{0.01\textwidth}
        \notsotiny{\rotatebox{90}{\hspace{.5cm}\textsf{relative time}}}
    \end{subfigure} 
    \begin{subfigure}[t]{.24\linewidth}
    \includegraphics[width=\linewidth]{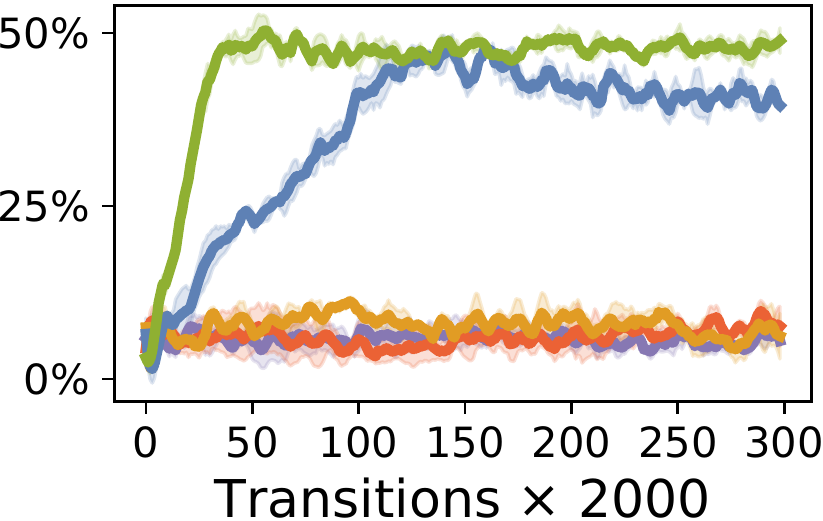}\vspace{-.3em}
    \caption{1 object moves}
    \end{subfigure}
    \hfill
    \begin{subfigure}[t]{.24\linewidth}
    \includegraphics[width=\linewidth]{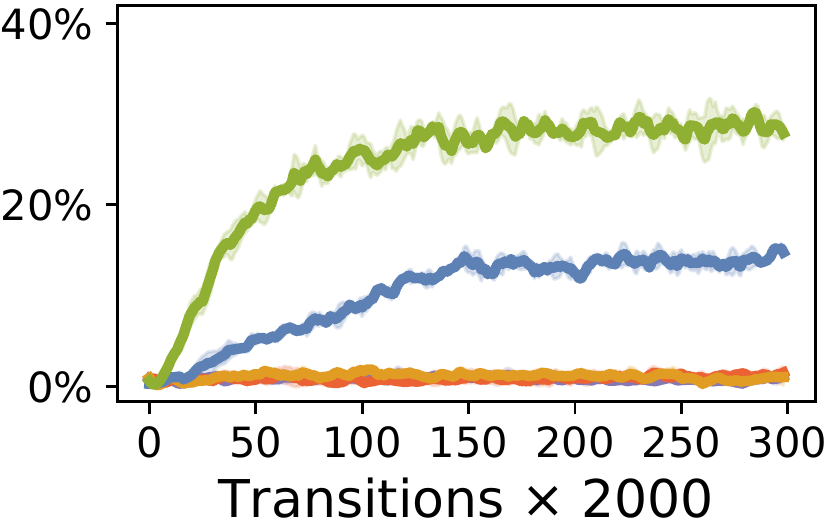}\vspace{-.3em}
    \caption{2 or more objects move}
    \end{subfigure}\hfill
    \begin{subfigure}[t]{.24\linewidth}
    \includegraphics[width=\linewidth]{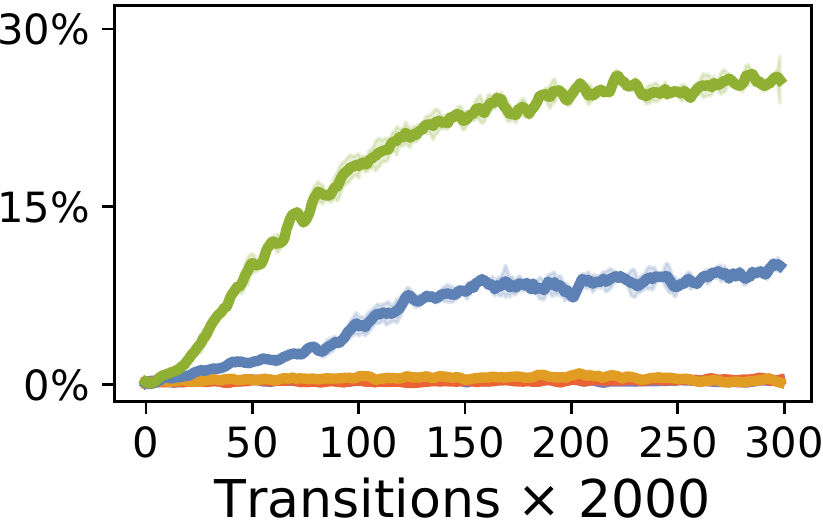}\vspace{-.3em}
    \caption{object(s) flipped}
    \end{subfigure}
    \begin{subfigure}[t]{.24\linewidth}
    \includegraphics[width=\linewidth]{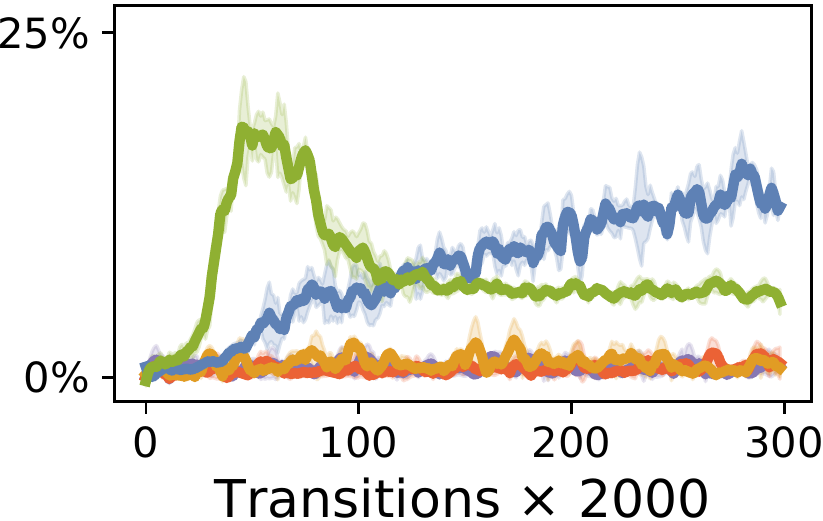}\vspace{-.3em}
    \caption{object(s) in air}
    \end{subfigure}
    \caption{{\bf Rich interactions during free play in \FPP by \ourMethod{}}.
     Interaction metrics of free-play exploration count the relative amount of time steps spent in moving one object (a), moving two and more objects (b), flipping object(s) (c), and moving objects into the air (d). \ourMethod{} engages quickly in all sorts of interactions with objects. The ablation with MLPs eventually performs similarly but is less sample efficient. The RL baselines perform poorly on these timescales. We used three independent seeds.
    }
    \label{fig:interactions:FPP}
\end{figure}

In the \FPP environment, the situation is even more drastic, as shown in \fig{fig:interactions:FPP}.
\ourMethod starts repeatedly moving one object after only eight iterations (each 2000 environment steps), picks up objects after around 30 iterations, and continues with throwing and flipping objects and moving multiple objects frequently after 50 iterations. 
In \fig{fig:emergent:behavior} exemplary behaviors are shown in terms of snapshots at different stages of learning. 
We believe this sample efficiency is remarkable. 
The ablation of our method with MLP models is also able to engage in interesting and diverse interactions, but at a much slower rate (\fig{fig:interactions:FPP}).
The policy learning baselines rarely interact with objects, even after 300 iterations with 20 episodes each. 

\subsection{Zero-Shot Generalization to Downstream Tasks}

\setcounter{figure}{4}

\begin{figure}[b!]
    \centering
    \begin{minipage}[c]{.43\linewidth}
    \begin{subfigure}[t]{.49\linewidth}
    \includegraphics[width=\linewidth]{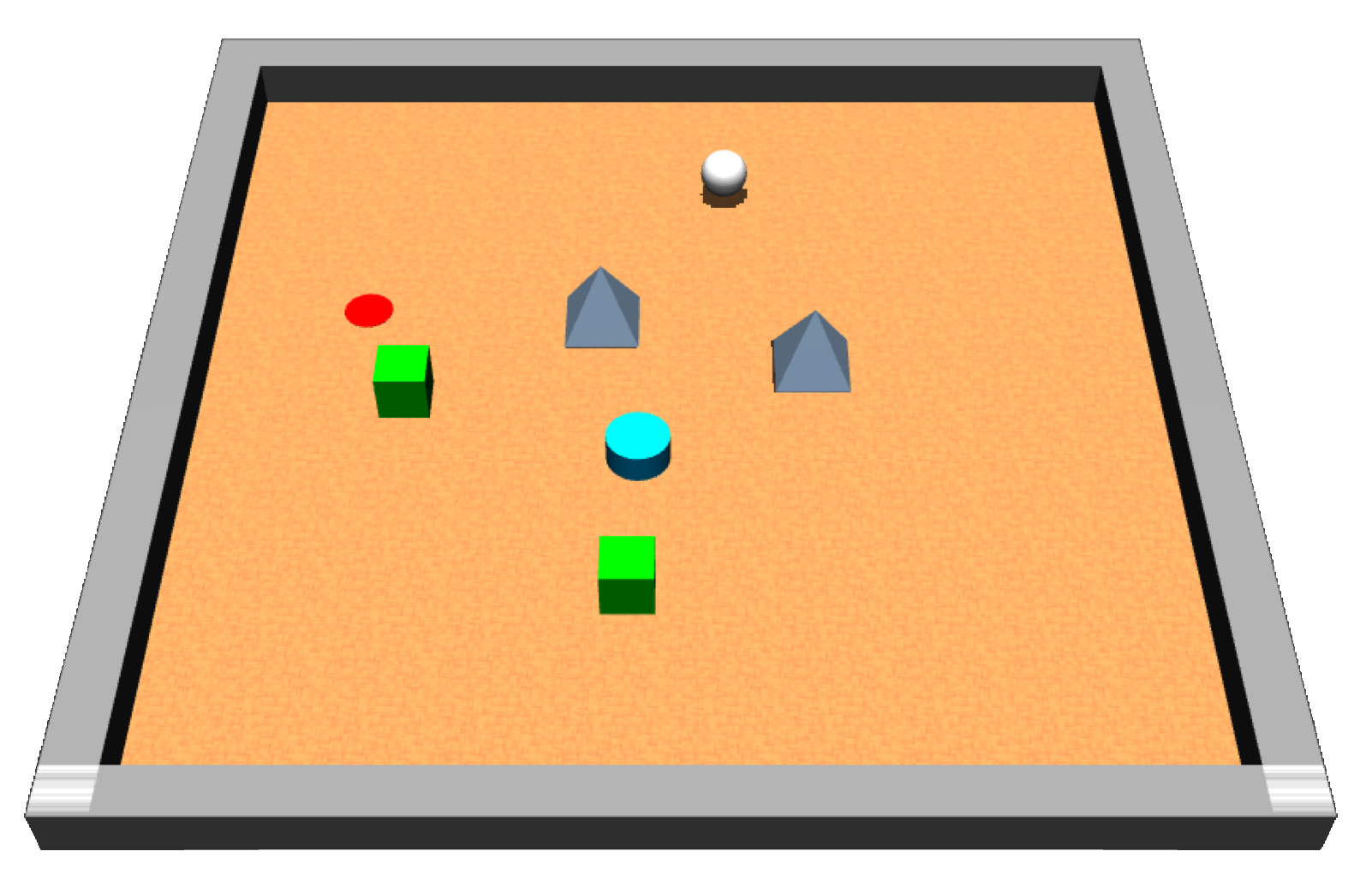}
    \caption{$t=0$}
    \end{subfigure}
    \hfill
    \begin{subfigure}[t]{.49\linewidth}
    \includegraphics[width=\linewidth]{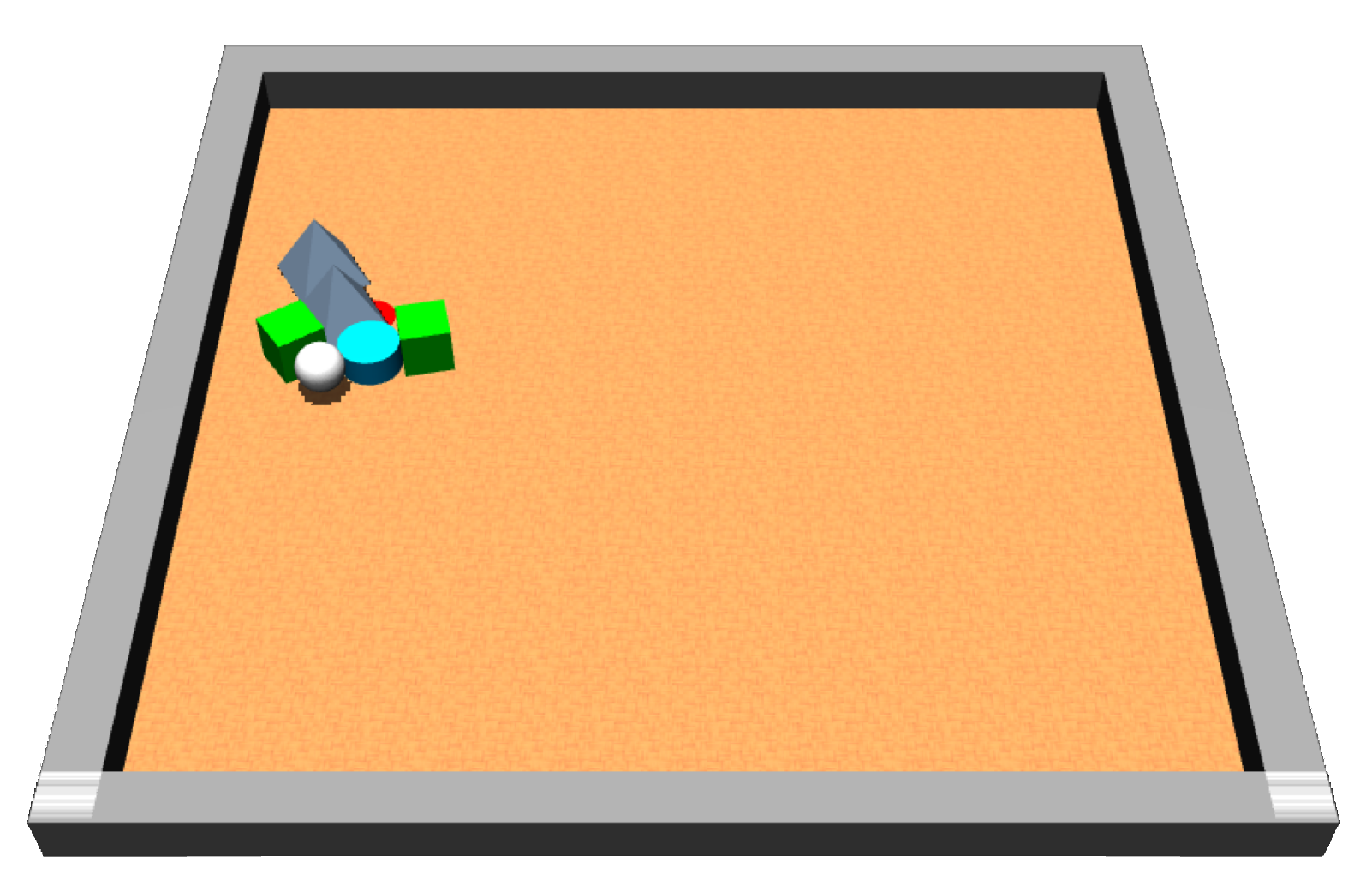}
    \caption{$t=500$}
    \end{subfigure}\vspace{-.5em}
    \captionof{figure}{Downstream task in \Playground: move all objects to one place, as solved by \ourMethod.}
    \label{fig:playground:task}
    \end{minipage}
    \hfill
    \begin{minipage}[c]{.535\linewidth}
  \captionof{table}{\textbf{Success rates in the \Playground environment} for the downstream task \textit{Push objects to one location}, with a variable number of objects after 100 and 250 training iterations of free play. Computed over three random seeds, each evaluated on 100 rollouts.}
  \vspace{-.3em}
  \label{tab:success_playground}
  \centering
  \renewcommand{\arraystretch}{1.03}
  \resizebox{.98\linewidth}{!}{ %
  \begin{tabular}{@{}rcccc@{}}
    \toprule
               & 4 Objs@100   & 4 Objs@250   & 3 Objs@250     & 5 Objs@250 \\
               & one each & one each &random     & random\\
    \midrule
    \ourMethod &  $0.70 \pm 0.04$ & $0.91\pm 0.06$   & $0.93\pm 0.03$   &  $0.93\pm 0.04$\\                                          
    \mlpicem   &  $0.08 \pm 0.01$ & $0.90\pm 0.03$    & ---   &  ---     \\
    \bottomrule
  \end{tabular}  
  }
    \end{minipage}
\end{figure}

We demonstrate that after the free-play phase, the learned models can be used for zero-shot solving of complex downstream tasks without the necessity to generate new data or to perform further training. 
Furthermore, thanks to the combinatorial generalization capabilities of GNNs, the model can be used with a different number of objects in the environment than seen during free play.
The only baseline we consider for the zero-shot generalization performance is \mlpicem, as none of the other methods discussed in the previous section can solve downstream tasks without additional training.

In \Playground, we consider the task of bringing all objects to a single target location, as shown in \fig{fig:playground:task}. 
The reward function is the sum of negative distances of all objects to the target outside a threshold distance. 
The target location is sampled randomly at the beginning of each episode. The success rates for different numbers of objects are shown in \tab{tab:success_playground}. 
We define success rate in the multi-object setup as the fraction of objects solved relative to the total number of objects spawned in the environment.
After 100 training iterations of free play, \ourMethod already achieves a 70\% success rate, whereas the MLP version only reaches 8\%. After 250 iterations, both methods are on par. Since \ourMethod can also deal with a variable input dimension, we also consider the task with more or less objects with randomly sampled types.

\setlength{\intextsep}{1.0pt plus 2.0pt minus 2.0pt}
\setlength{\columnsep}{10pt}%
\begin{wrapfigure}{r}{0.50\textwidth}
    \centering
    \includegraphics[width=\linewidth]{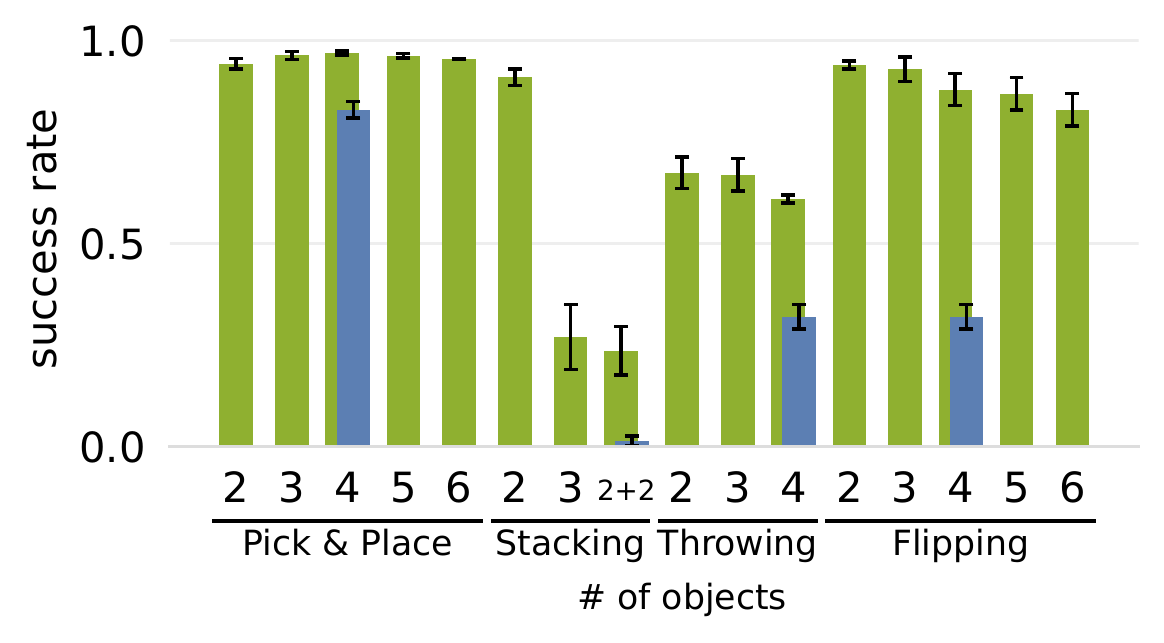}\vspace{-.8em}
    \caption{{\bf Zero-shot performance on downstream tasks in \FPP} for {\color{ourgreen}\ourMethod} 
    and {\color{ourblue}\mlpicem}.}
    \label{fig:fpp:performance}
\end{wrapfigure}
In \FPP, several challenging manipulation tasks need to be solved: pick \& place, stacking, throwing, and flipping, as shown in \fig{fig:overview}. The reward functions are detailed in \supp{app:rewards}. 
In \fig{fig:fpp:performance}, we present the success rates in each of the tasks for \ourMethod and the MLP-based planning baseline \mlpicem, where applicable.
We find remarkable success rates across the board for \ourMethod, even in challenging tasks such as stacking and throwing. 
For stacking, we report a success rate of 1 if the tower with all the objects in the environment is stacked and 0 otherwise. For the multi-tower task, denoted by $2+2$, we consider success 1 only when two towers are stacked.
We recommend visiting our website\footnote{\url{https://martius-lab.github.io/cee-us}} for videos of these tasks.
In \supp{app:model_rnd}, we present additional experiments combining RND as intrinsic reward with model-based planning, similar to \citet{Lambert2022:iMPC}, in \FPP, where we look at both structured and unstructured world models. 
The results showcase the benefits of using the model's own epistemic uncertainty estimate to guide exploration, as it is the case with ensemble disagreement,
leading to more accurate and more robust dynamics models and better zero-shot downstream task generalization. %

\subsection{Offline Learning of Downstream Tasks from Exploratory Data}\label{sec:offline} 

This experiment investigates the quality of the data collected by the different methods in the free-play phase for solving downstream tasks via offline RL.
This allows us to compare to other baselines that do not support zero-shot generalization.
Additionally, it is one way of obtaining task-specific policies.
In offline RL, a control policy is learned from a fixed dataset, where we repurpose the exploration data. 
The results are shown in \tab{tab:offline_rl:performance}. Offline policy learning is performed via CQL~\citep{kumar2020conservative}\footnote{We are using the implementation from the \textit{d3rlpy} library~\citep{seno2021d3rlpy}} and training details can be found in \supp{app:offline_rl}.
To learn different downstream tasks from the same dataset, reward relabeling and hindsight experience replay~\citep{NIPS2017_453fadbd} are used.

In accordance with the interaction metrics during the intrinsic phase (\figs{fig:interactions:playground}{fig:interactions:FPP}), \ourMethod and \mlpicem achieve the highest performance in all four tasks, supporting the benefit of optimizing for long-horizon future novelty.
Especially in the object manipulation tasks, data from \ourMethod{} shows a clear advantage. 
Tasks with more objects could not be solved with the amount of data collected.

\begin{table}[htb]
    \centering
    \caption{{\bf Success rates on tasks with offline RL} trained from exploration data generated during the intrinsic phase with the respective method.
    Using CQL and HER on the data for each task, we see that \ourMethod creates much more useful data for offline learning.
    }\vspace{.3em}
    \label{tab:offline_rl:performance}
    \renewcommand{\arraystretch}{1.03}
    \resizebox{.98\linewidth}{!}{ %
    \begin{tabular}{@{}cc|c@{\ \ }c@{\ \ }ccc@{}}
    \toprule
    Domain & Task & Disagreement~\cite{pathak2019self} & RND~\cite{burda2018exploration} & ICM~\cite{pathak2017curiosity} & \mlpicem & \ourMethod{}\\
\midrule
\Playground & Locomotion & $0.05 \pm 0.01$ & $0.35 \pm 0.05$ & $0.01 \pm 0.0$ & $1.0 \pm 0.0$ & $1.0 \pm 0.0$\\
500k datapoints & Relocate 1 obj. & $0.06 \pm 0.01$ & $0.06 \pm 0.0$ & $0.06 \pm 0.01$ & $0.2 \pm 0.06$ & $0.9 \pm 0.01$\\
\midrule
\FPP & Reach & $0.09 \pm 0.01$ & $0.19 \pm 0.05$ & $0.2 \pm 0.03$ & $0.65 \pm 0.09$ & $0.94 \pm 0.04$\\
600k datapoints & Pick \& Place 1 obj. & $0.07 \pm 0.0$ & $0.07 \pm 0.0$ & $0.07 \pm 0.01$ & $0.18 \pm 0.06$ & $0.43 \pm 0.07$\\
\bottomrule
    \end{tabular}
    }
\end{table}

\section{Related Work}\label{sec:related} 

\paragraph{Intrinsic motivation in RL} Prediction error \cite{schmidhuber1991possibility,pathak2017curiosity,Kim20:Active}, novelty and Bayesian surprise \cite{Storck1995Reinforcement-Driven-Information,BlaesVlastelicaZhuMartius2019:CWYC,PaoloEtAl2021:NoveltySearchSparseReward}, learning progress \cite{schmidhuber1991possibility, Colas2019:CURIOUS, BlaesVlastelicaZhuMartius2019:CWYC}, empowerment \cite{KlyubinPolani2005:Empowerment,mohamed2015variational} and count-based metrics \cite{burda2018exploration,Lambert2022:iMPC} are among the popular intrinsic reward signal definitions used in RL. These intrinsic rewards are either used to aid exploration in challenging tasks with sparse task rewards or in a task-agnostic setup where they are the only rewards. %
Algorithms using the task-agnostic setting follow two routes to solve downstream tasks:  
(i) they re-label the collected data during intrinsic exploration with the downstream task reward and perform offline RL \cite{sekar2020planning,yarats2022don,Lambert2022:iMPC} or 
(ii) they use snapshots of the exploration policy for bootstrapping and fine-tune them on downstream tasks \cite{groth2021curiosity,laskin2021urlb}. 
The first variant does not require additional interaction with the environment, however, additional training of a new policy is still necessary. In addition, offline RL struggles with the distribution shift, while we show that a good world model can generalize better.
The second variant suffers from another inherent issue: the emerging behaviors of the exploration policy are lost during training. 
Although \citet{groth2021curiosity} address this problem by snapshotting and performing hierarchical RL, this is more of a bandaid solution. 
Many intrinsically motivated learning systems use a goal-achieving setting with predefined goal spaces \cite{Colas2019:CURIOUS,BlaesVlastelicaZhuMartius2019:CWYC,openai2021:assymmetric-selfplay,Mendonca2021:LEXA} or auxiliary tasks \cite{Riedmiller2018:LearningByPlaying}.
The differences in the existing paradigms are highlighted in \tab{tab:overview}. Methods such as ICM \cite{pathak2017curiosity} rely on retrospective intrinsic motivation (here prediction error), \ie the agent has to already be in the next state to assess how novel that state is. In order to accommodate planning for multi-step intrinsic motivation signals into the future, we need a way to predict them. \citet{Huber18:Emergent}, for instance, do so by learning a loss model to predict the prediction error at future time steps. 
Another line of research, including our method, uses the disagreement of an ensemble of world models to estimate the predictive information gain as the intrinsic motivation measure \cite{pathak2019self, sekar2020planning}.
Plan2Explore \cite{sekar2020planning} shares similarities to our approach in the task-agnostic exploration phase, using multi-step ensemble disagreement as intrinsic reward. It works with a latent dynamics prediction model and has been applied to domains with image observations. However, Plan2Explore does not use structured world models and lacks mechanisms to achieve combinatorial generalization that is beneficial for sample-efficient exploration in object manipulation tasks.

\paragraph{Relational Networks} 
Several works showcase improved dynamics prediction performance in environments with interacting entities using structured world models \cite{kipf2019contrastive,sanchez2020learning,watters_NIPS2017}. 
\citet{kipf2019contrastive} uses a GNN to learn the latent transition dynamics in simple manipulation tasks with 2D shapes and 3D blocks from images. However, they use an object-factorized action space and do not tackle exploration but rely on an external dataset. 
\citet{sanchez2020learning} applies GNNs to learn the dynamics of physical bodies, where the entities correspond to joints, also without active exploration.
\citet{Driess2022:Nerf} uses GNNs and NERFs to obtain object-centric representations from images and RRT planners.
\citet{biza2022factored} also utilizes the combinatorial generalization of structured world models to achieve zero-shot task generalization. However, the policy uses parameterized high-level actions and the world model is learned on an offline expert dataset, thus sidestepping the exploration challenge.
Outside the model-based paradigm, \citet{li2020towards} achieves block stacking in \FPP using a GNN policy, attention and tailored learning curriculum.
In contrast, we use GNNs for computing intrinsic rewards and online planning.  

\paragraph{Curiosity in Object-Centric RL}
\citet{watters2019cobra} deploy curious exploration with retrospective prediction error as an intrinsic reward in an object-centric setting with image inputs and a pre-trained vision model for object discovery, but without object-object interactions. %
In \citet{Seitzer2021CID}, an object-centric causal action-influence is used as an intrinsic reward to improve sample efficiency.

\section{Discussion} \label{sec:discussion}

In this work, we present \ourMethod that combines the learning of GNNs as structured world models with curiosity-driven, planning-based exploration. We tackle a significant challenge that is not often addressed in existing intrinsically motivated RL works: guiding exploration towards potentially \textbf{useful} components of the environment. By approximating information gain via world models injected with relational inductive biases, \ourMethod focuses on the interactions between entities in the environment. Maximizing future information gain via multistep forward-planning enables \ourMethod{} to discover interaction-rich behaviors more efficiently than the exploration-policy-based baselines, as well as planning-based paradigms without structured world models. 

After the intrinsic free-play phase, we use the learned GNNs to solve downstream tasks with model-based planning and zero-shot without any additional training.  
\ourMethod can achieve zero-shot generalization without the need for an additional policy learning step, a strength of \ourMethod{} that is missing from all the baselines. We show that even non-structured MLPs achieve zero-shot generalization on some tasks, albeit with lower success rates. This indicates that there are benefits to learning good world models and utilizing them for control even without any further structural biases. This also ties into knowledge-based intrinsic motivation, as the experience gathered during free play is distilled and stored in the learned world models instead of being dismissed. 
It is important to note that using model-based planning during the intrinsic phase does not restrict us to model-based approaches in the extrinsic phase. 
The learned model could also be used to extract a task policy later using model-based policy optimization~\cite{janner2019trust} or to perform offline RL on the data generated during free play, which we demonstrate in \sec{sec:offline}. 

Despite the sample efficiency, there are some limitations to model-based online planning. The complexity of behaviors discovered in free play is upper-bounded by the finite planning horizon. The same also applies to the extrinsic phase, where solving tasks like throwing objects or solving multistep manipulation tasks require longer planning horizons without extensive reward shaping.

Although \ourMethod currently uses proprioceptive state information, we do not assume access to any privileged information such that there are no fundamental limitations prohibiting us from applying it to real robots. Following \citet{kipf2019contrastive} and \citet{watters2019cobra}, \ourMethod can also be extended to deal with image inputs and use methods like SCALOR \cite{JiangJanghorbaniDeMeloAhn2020SCALOR} to extract object-centric unsupervised representations. The differentiation between agent and objects can be identified unsupervised \cite{Zadaianchuk22:SMORL}.  

The demonstrated sample efficiency in the unsupervised learning of capable world models opens new avenues for learning directly on real hardware. To put the training time into context, 45 training iterations correspond to 1h of interaction time. So our downstream performance in \FPP was achieved after about 6.5h of free-play from scratch.

\begin{ack}
The authors thank Arash Tavakoli and Pavel Kolev for helpful discussions and Andrii Zadaianchuk, Christian Gumbsch and Nico Gürtler for their help reviewing the manuscript. 
The authors thank the International Max Planck Research School for Intelligent Systems (IMPRS-IS) for supporting Cansu Sancaktar and Sebastian Blaes.
Georg Martius is a member of the Machine Learning Cluster of Excellence, EXC number 2064/1 -- Project number 390727645. 
We acknowledge the financial support from the German Federal Ministry of Education and Research (BMBF) through the Tübingen AI Center (FKZ: 01IS18039B).
This work was supported by the Volkswagen Stiftung (No 98 571).
\end{ack}

\nocite{kannan2021robodesk}
\setlength{\bibsep}{3.7pt}  %
\bibliography{neurips_2022}

\section*{Checklist}

\begin{enumerate}

\item For all authors...
\begin{enumerate}
  \item Do the main claims made in the abstract and introduction accurately reflect the paper's contributions and scope?
    \answerYes{}
  \item Did you describe the limitations of your work?
    \answerYes{We discuss the limitations of our work in \sec{sec:discussion}.}
  \item Did you discuss any potential negative societal impacts of your work?
    \answerNA{We believer there are no negative societal impacts of our work.}
  \item Have you read the ethics review guidelines and ensured that your paper conforms to them?
    \answerYes{}
\end{enumerate}

\item If you are including theoretical results...
\begin{enumerate}
  \item Did you state the full set of assumptions of all theoretical results?
    \answerNA{}
        \item Did you include complete proofs of all theoretical results?
    \answerNA{}
\end{enumerate}

\item If you ran experiments...
\begin{enumerate}
  \item Did you include the code, data, and instructions needed to reproduce the main experimental results (either in the supplemental material or as a URL)?
    \answerYes{We provide the algorithmic details of our method in \sec{sec:algorithm} and further details in supplementary material. The code is published on the project website.}
  \item Did you specify all the training details (e.g., data splits, hyperparameters, how they were chosen)?
    \answerYes{See \supp{app:hyperparams}}
        \item Did you report error bars (e.g., with respect to the random seed after running experiments multiple times)?
    \answerYes{}
        \item Did you include the total amount of compute and the type of resources used (e.g., type of GPUs, internal cluster, or cloud provider)?
    \answerYes{Further details are in the \supp{app:hyperparams}.}
\end{enumerate}

\item If you are using existing assets (e.g., code, data, models) or curating/releasing new assets...
\begin{enumerate}
  \item If your work uses existing assets, did you cite the creators?
    \answerYes{}
  \item Did you mention the license of the assets?
    \answerNA{}
  \item Did you include any new assets either in the supplemental material or as a URL?
    \answerYes{Code and videos are published in the project website.}
  \item Did you discuss whether and how consent was obtained from people whose data you're using/curating?
    \answerNA{}
  \item Did you discuss whether the data you are using/curating contains personally identifiable information or offensive content?
    \answerNA{}
\end{enumerate}

\item If you used crowdsourcing or conducted research with human subjects...
\begin{enumerate}
  \item Did you include the full text of instructions given to participants and screenshots, if applicable?
    \answerNA{}
  \item Did you describe any potential participant risks, with links to Institutional Review Board (IRB) approvals, if applicable?
    \answerNA{}
  \item Did you include the estimated hourly wage paid to participants and the total amount spent on participant compensation?
    \answerNA{}
\end{enumerate}

\end{enumerate}

\newpage

\appendix

\renewcommand{\thefigure}{S\arabic{figure}}
\renewcommand{\thetable}{S\arabic{table}}
\renewcommand{\thealgorithm}{S\arabic{algorithm}}
\renewcommand{\theequation}{S\arabic{equation}}

\newcommand{\gnnrndicem}{\mbox{GNN\,+\,RND}\xspace}
\newcommand{\mlprndicem}{\mbox{MLP\,+\,RND}\xspace}

\setcounter{figure}{0}
\setcounter{table}{0}
\setcounter{algorithm}{0}
\setcounter{equation}{0}

\begin{center}\Large \textbf{Supplementary Material for \\
Curious Exploration via Structured World Models Yields Zero-Shot Object Manipulation}\end{center}

\section{GNN Architectural Details} \label{app:gnn_architecture}
We use message-passing GNNs in \ourMethod as described in \sec{sec:gnn}.
When different object types are present in the environment, we also include static object features in the object states $s_t^i$. The dynamic component $s_t^{\text{dyn},i}$ is time-dependent and contains object positions and velocities,
whereas the time-independent static features $s^{\text{stat},i}$ contain identifiers for different object types. For the overall object state at time step $t$,
we get $s_t^i = [s_t^{\text{dyn},i}, s^{\text{stat},i}]$. This can be viewed as a concatenation of a dynamic and a static graph \cite{sanchez2020learning}. 
The GNN only makes a next state prediction for the dynamic state component of the object nodes so that \eg the node update is given by:
\begin{align}
    \hat{s}_{t+1}^{\text{dyn},i} &= g_{\text{node}}\big(\big[s_t^{\text{dyn},i}, s^{\text{stat},i}, c, \operatorname{aggr}_{i \neq j}\big(e_t^{(i,j)}\big)\big]\big).
\end{align}

We use static features only in the \Playground environment, where we have 4 different object types (cube, light cube, cylinder and pyramid) and the color of each object is used as the static feature.
The mean is used as the permutation-invariant aggregation function $\operatorname{aggr}$.

\section{Planning Details}\label{app:icem}
For planning, we use the improved Cross-Entropy Method (iCEM)~\cite{pinneri2020:iCEM}. The pseudocode is given in \alg{alg:improved-cem}. The costs for the planner correspond to negative reward such that: 
\begin{equation*}
C(s_t, a_t, s_{t+1}) = -R(s_t, a_t, s_{t+1}),
\end{equation*}
where $R$ can be intrinsic rewards $R_I$ or extrinsic task rewards $R_\text{task}$. Whenever we are dealing with an ensemble of models, we use the same notation $R(s_t, a_t, s_{t+1})$, even though the reward function in this case takes the transitions of the whole ensemble $\{(s_t^m, a_t , s_{t+1}^m)\}_{m=1}^M$ as input arguments. \footnote{Note that we overload the superscript to both indicate ensemble members' predictions and object-centric state representations. The index $m$ is for the the prediction of ensemble member $m$ on the whole state, and $i$ signals that we are looking at the state of object $i$.}

The algorithm shown here is for a single model $f_{\theta}$. In the case of an ensemble of models $\{({f_{\theta}}_m)_{m=1}^M\}$ with ensemble size $M$, each model sees the same P sampled action trajectories with each \( \mathbf{a}_t = (a_{t+h})_{h=0}^{H-1} \in \mathbb{R}^{n_a \times H}\).
During the intrinsic phase of \ourMethod, the intrinsic rewards for planning are computed based on the ensemble disagreement, such that $R_I(s_t, a_t, s_{t+1})$ for one time step in a simulated trajectory is a scalar, computed according to \eqn{eqn:epistemic}. 

In the extrinsic phase when we have task-specific reward functions, we utilize the different ensemble predictions for more robust action selection. Each model of the ensemble creates a cost trajectory for each sampled action sequence with $\{R_\text{task}(s_{t+h}, a_{t+h}, s_{t+h+1})\}_{h=0}^{H-1}$ such that the overall cost of a sampled trajectory amounts to a tensor with size $M \times H$.
In order to then select the elites, we average the costs over the ensembles.

In a generalized setup, the sum in \eqn{eqn:action_planner} can be replaced with another permutation-invariant function $\phi$. For planning, other than the default mode \texttt{sum} as shown in equation \eqn{eqn:action_planner}, we also allow mode \texttt{best} with $\phi=\max\left(\{R(s_{t+h}, a_{t+h}, s_{t+h+1})\}_{h=0}^{H-1}\right)$, which chooses the optimal trajectory according to the best reward observed at any time step over the planning horizon.

\begin{algorithm}
  \caption{Model predictive control with iCEM planner}
 \label{alg:improved-cem}
 \begin{algorithmic}[1]
  \State \textbf{Input:} $P$: number of samples; $H$: planning horizon; $n_a$: action dimension; $K$: size of elite-set; $\beta$: colored-noise exponent, \textit{CEM-iterations}: number of iterations; $\xi$: fraction of elites reused; $\sigma_{init}$: noise strength, $\alpha$: momentum,
  $f_{\theta}$: Forward dynamics model, $\phi$: permutation-invariant function to compute overall cost over planning horizon.
  \For{t = 0 \textbf{to} T$-1$} \Comment{loop over episode length}
    \If{t == 0}
      \State $\mu_0$ $\gets$ constant vector in $\mathbb{R}^{n_a\times H}$
    \ElsIf{\texttt{shift\_elites}}
      \State $\mu_t$ $\gets$ shifted $\mu_{t-1}$ (and repeat last time-step) %
    \EndIf
    \State $\sigma_t$ $\gets$ constant vector in $\mathbb{R}^{n_a\times H}$ with values $\sigma_{init}$

    \For{i = 0 \textbf{to} CEM-iterations$-1$}

      \State samples $\gets$ $P$ samples from clip$(\mu_t + \mathcal{C}^\beta(n_a, H) \odot \sigma_t^2)$

      \If{i == 0 \textbf{and} \texttt{shift\_elites}}
        \State add fraction $\xi$ of \textbf{shifted} elite-set$_{t-1}$ to samples
      \ElsIf{\texttt{keep\_elites}}
        \State add fraction $\xi$ of elite-set$_t$ to samples
      \EndIf

      \If{ i == last-iter}
         \State add mean to samples
      \EndIf

      \For{h = 0 \textbf{to} H-1} \Comment{Compute state trajectories in model $f_{\theta}$'s imagination}
        \State $s_{t+h+1} \gets f_{\theta}(s_{t+h}, a_{t+h})$ for $a_{t+h}$ in samples
      \EndFor
      \State rewards $\gets$ rewards of sampled trajectories $\phi(\{R(s_{t+h}, a_{t+h}, s_{t+h+1})\}_{h=0}^{H-1})$ %

      \State elite-set$_t$ $\gets$ best $K$ samples according to rewards

      \State $\mu_t$, $\sigma_t$ $\gets$ fit Gaussian distribution to elite-set$_t$ with momentum $\alpha$
    \EndFor
    \If{\texttt{use\_mean\_actions}}
        \State execute first action of mean of elite sequences
    \Else
        \State execute first action of best elite sequence
    \EndIf
  \EndFor
 \end{algorithmic}

\end{algorithm}

\section{Experiment Details} \label{app:hyperparams}
In this section, we provide experimental details and hyperparameter settings.
\subsection{Intrinsic Phase with \ourMethod}\label{app:intrinsic}
In the intrinsic phase of \ourMethod, we iteratively generate rollouts with the iCEM planner using intrinsic rewards and then train the models of the ensemble on the overall data collected so far. We test \ourMethod on the \Playground and \FPP environments. The environment properties as well as the episode lengths and model training frequencies are given in \tab{tab:environment_parameters}. Four objects are present in each environment during free play (in \Playground: one of each object type). The parameters for the GNN model architecture as well as the training parameters for model learning are listed in \tab{tab:gnn_model_training_settings}. Note that model learning only occurs during the intrinsic phase. For the extrinsic phase, we take the learned model with the listed architectural settings to solve downstream tasks zero-shot.

The intrinsic free-play in \ourMethod, together with the data collection and consequent model updates, is run for 300 training iterations in \FPP, which takes roughly 72 hours using a single GPU (here NVIDIA GeForce RTX 3060) and 6 cores on an AMD Ryzen 9 5900X Processor. In \Playground, we run \ourMethod for 250 training iterations which takes \circa 50 hours. Note that the duration of one training iteration increases throughout free-play, since we train the model for a fixed number of epochs on the whole data collected so far. As the number of transitions stored in the buffer increases, the number of update steps for the same number of epochs also increases. 

\begin{table*}[htb]
    \centering
    \caption{Environment settings. In both environments 2000 transitions are generated within one training iteration of \ourMethod.}
    \label{tab:environment_parameters}
    \begin{subtable}[t]{.5\textwidth}
        \centering
       \begin{tabular}{@{}ll@{}}
        \toprule
        \multicolumn{2}{c}{\Playground} \\
        \textbf{Parameter} & \textbf{Value} \\
        \midrule
        Episode Length & $200$ \\
        Train Model Every & $10$ Episodes \\
        Action Dim. & $2$ \\
        Robot/Agent State Dim. & $4$ \\
        Object Dynamic State Dim. & $6$ \\
        Object Static State Dim. & $3$ \\
        \bottomrule
        \end{tabular}
    \end{subtable}%
    \begin{subtable}[t]{.5\textwidth}
        \centering
        \begin{tabular}{@{}ll@{}}
        \toprule
        \multicolumn{2}{c}{\FPP} \\
        \textbf{Parameter} & \textbf{Value} \\
        \midrule
        Episode Length & $100$ \\
        Train Model Every & $20$ Episodes \\
        Action Dim. & $4$ \\
        Robot/Agent State Dim. & $10$ \\
        Object Dynamic State Dim. & $12$ \\
        Object Static State Dim. & $0$ \\
        \bottomrule
        \end{tabular}
    \end{subtable}
\end{table*}

\begin{table*}[htb]
    \centering
    \caption{Base settings for GNN model training in intrinsic phase of \ourMethod.}
    \label{tab:gnn_model_training_settings}
    \begin{subtable}[t]{.5\textwidth}
        \centering
        \caption{General settings.}
        \begin{tabular}{@{}ll@{}}
        \toprule
        \textbf{Parameter} & \textbf{Value} \\
        \midrule
        Network Size of $g_\text{node}$ & $2 \times 128$ \\
        Network Size of $g_\text{edge}$ & $2 \times 128$ \\
        Network Size of $g_\text{global}$ & $2 \times 128$ \\
        Activation function & ReLU \\
        Layer Normalization & Yes\\
        Number of Message-Passing & 1\\
        Ensemble Size & 5\\
        Optimizer & ADAM \\
        Batch Size & $125$ \\
        Epochs & $25$ \\
        Learning Rate & $10^{-5}$ \\
        Weight Decay & $0.001$ \\
        Weight Initialization & Truncated Normal\\
        Normalize Input & Yes \\
        Normalize Output & Yes \\
        Predict Delta & Yes \\
        \bottomrule
        \end{tabular}
    \end{subtable}%
    \begin{subtable}[t]{.5\textwidth}
        \centering
        \caption{Environment-specific settings.}
        \begin{tabular}{@{}ll@{}}
        \toprule
        \multicolumn{2}{c}{\Playground} \\
        \textbf{Parameter} & \textbf{Value} \\
        \midrule
        Network Size of $g_\text{node}$ & $2 \times 64$ \\
        Network Size of $g_\text{edge}$ & $2 \times 64$ \\
        Network Size of $g_\text{global}$ & $2 \times 64$ \\
        Learning Rate & $0.0001$ \\
        Weight decay & $5 \cdot 10^{-5}$ \\
        \bottomrule
        \\
        \toprule
        \multicolumn{2}{c}{\FPP} \\
        \textbf{Parameter} & \textbf{Value} \\
        \midrule
        \multicolumn{2}{c}{Same as general settings} \\
        \bottomrule
        \end{tabular}
    \end{subtable}
\end{table*}

\subsection{Controller Parameters}
The set of default hyperparameters used for the iCEM controller are presented in \tab{tab:controller_settings}, as well as environment-specific controller settings used for the intrinsic phase of \ourMethod.

\begin{table*}[htb]
    \centering
    \caption{Base settings for iCEM in \ourMethod as well as the environment-specific settings used in the intrinsic phase. Same settings are used for \mlpicem.}
    \label{tab:controller_settings}
    \begin{subtable}[t]{.48\textwidth}
        \centering
        \caption{General settings.}
        \begin{tabular}{@{}ll@{}}
        \toprule
        \textbf{Parameter} & \textbf{Value} \\
        \midrule
        Number of samples $P$ & $128$ \\
        Horizon $H$ & $30$ \\
        Size of elite-set $K$ & $10$ \\
        Colored-noise exponent $\beta$ & $3.5$ \\
        \textit{CEM-iterations} & $3$ \\
        Noise strength $\sigma_{\text{init}}$ & $0.5$ \\
        Momentum $\alpha$ & $0.1$ \\
        \texttt{use\_mean\_actions} & Yes \\
        \texttt{shift\_elites} & Yes \\
        \texttt{keep\_elites} & Yes \\
        Fraction of elites reused $\xi$ & $0.3$ \\
        Cost along trajectory & \texttt{sum} \\
        \bottomrule
        \end{tabular}
    \end{subtable}%
    \begin{subtable}[t]{.48\textwidth}
        \centering
        \caption{Environment-specific settings.}
        \begin{tabular}{@{}ll@{}}
        \toprule
        \multicolumn{2}{c}{\Playground} \\
        \multicolumn{2}{c}{\emph{Intrinsic Phase}} \\
        \textbf{Parameter} & \textbf{Value} \\
        \midrule
        \texttt{shift\_elites} & No \\
        \texttt{keep\_elites} & No \\
        Noise strength $\sigma_{\text{init}}$ & $0.8$ \\
        \bottomrule
        \\
        \toprule
        \multicolumn{2}{c}{\FPP} \\
        \multicolumn{2}{c}{\emph{Intrinsic Phase}} \\
        \textbf{Parameter} & \textbf{Value} \\
        \midrule
        \multicolumn{2}{c}{Same as general settings} \\
        \bottomrule
        \end{tabular}
    \end{subtable}
\end{table*}

\subsection{Extrinsic Phase}\label{app:extrinsic}
In this section, we provide details on the extrinsic phase of \ourMethod, where the learned GNN ensemble is used to solve downstream tasks zero-shot via model-based planning. 

\subsubsection{Details on Downstream Tasks and Reward Functions}\label{app:rewards}
We use the notation introduced in \sec{sec:preliminaries}, where $s^i$ for $i=1, \ldots, N$ denotes the state of each of the $N$ objects present in the environment and $g^i$ denotes the goal for each object in the environment. The superscript $i$ is omitted for the goal if all objects' goals are the same. For ease of notation in the reward function definitions, we consider $s^i$ to be the achieved goal state, which for all tasks other than flipping corresponds to the positional information of each object's state (x-y for \Playground and x-y-z for \FPP). The actuated agent, \ie robot, state is given by $s^\text{agent}$. Unless stated otherwise, the L2-norm is used to compute the distance between the current state $s$ and a target/goal state $g$ denoted by $\text{dist}(s,g) = \norm{s-g}_2$. We use $\delta$ to denote the environment threshold for goal distances used to compute sparse rewards as well as potential cut-off values for dense rewards. 
The sizes in both environments are on different scales, so the used $\delta$ values vary. In \Playground, the spherical agent's diameter is $0.2$ and objects have size ca. $0.2$ with slight variations. 
In \FPP, each cube/block has size 0.05.

\paragraph{\Playground-Pushing}
The task in the \Playground environment is defined as bringing all objects to a target location $g \in \mathbb{R}^2$ that is sampled randomly in the beginning of each episode. The reward is defined as the sum of the negative distances of each object to the target location up to a threshold distance $\delta$ such that:
\begin{equation}
    R_\text{push} = \sum_{i=1}^N -\max(\text{dist}(s^i, g), \delta).
\end{equation}
The reason we have a cut-off at distance $\delta$ is to ensure that the agent doesn't unnecessarily try to bring each object to the exact center since the task is to bring all objects to the target location and the overall reward cannot be zero in the end with more than one object in the environment. We still use a small $\delta$ of 0.23 for this buffer zone in the experiments, such that the model still has to find a plan that focuses on the next unsolved object instead of optimizing for small gains with an object that is essentially already at the target $g$. Note that for evaluating whether the task was successfully solved, we define the distance threshold to be larger $\delta_\text{eval}=0.35$, as to accurately account for cases where the target is in the corners or at the wall, such that not all objects can fit in the buffer zone area with the conservative $\delta$.

\paragraph{\FPP-Stacking} For stacking, sparse incremental rewards are used with reward shaping. The shaped reward contains the (dense) distance between the gripper and the position of the next block to be stacked in the tower given by $s^\text{next}$. If the tower is fully stacked, then the shaped reward component $s^\text{next}$ contains the distance of the gripper to a resting position away from the tower base. In the experiments, we use $(0,0,0)$ as the goal position for the robot after it has finished stacking. The reward function is given by:
\begin{equation}
    R_\text{stack} = \sum_{i=1}^N \left(-1+\llbracket \text{dist}(s^i, g^i)<\delta \rrbracket \right) - \eta \cdot \text{dist}(s^\text{agent}, s^\text{next}),
\end{equation}
where $\llbracket \cdot \rrbracket$ are Iverson brackets and $\eta$ is the scale of the shaped reward component. In the experiments, we use $\delta=0.02$ (each block has size $0.05$) and $\eta=0.01$. Note that in the original environment proposed in \citet{li2020towards}, the distance threshold for all tasks was defined to be 0.05. However, in order to ensure stable stacking we reduce this to 0.02 and also use this same value for evaluating for a successful stack. We do not allow any mismatch between $\delta$ and $\delta_\text{eval}$, when we have sparse rewards.

\paragraph{\FPP-Pick \& Place}
The task is defined as bringing each object $i$ to its individual goal position $g_i$ that is randomly sampled. In an environment with $N$ objects, the first $N-1$ have goal positions on the ground. The $N$th object's goal is in air with 50\% probability, where the target height is also sampled randomly.
We use a dense reward with the sum of the negative distances of each object position $s^i$ to its individual goal position $g^i$
\begin{equation}
    R_\text{pp} = \sum_{i=1}^N -\max(\text{dist}(s^i, g^i), \delta),
\end{equation}
with $\delta=0.025$. Similar to the \Playground-Push task, we use the $\delta$ value in the reward to ensure that the model doesn't over-optimize for each object. This is again set to be a more conservative distance threshold than the evaluation threshold $\delta_\text{eval}=0.05$, that is set to be the same as in the original environment \cite{li2020towards}.

\paragraph{\FPP-Throwing}
The task is to throw blocks onto goal sites with size 0.2 by 0.2 (so 4 times the size of each block). Each goal site is at least 0.16 and at most 0.20 away from the manipulability range of the robot arm. For this task, we use sparse rewards together with a dense component.  Throwing is a challenging task as (i) goal locations for objects are farther away, requiring a longer planning horizon and (ii) in the case of planning with dense rewards the agent can easily be stuck in a local optimum and push blocks outside of its manipulability range without actually reaching the goal location. In order to deal with (i), we keep a dense component for the reward and to address (ii), we also include sparse rewards and use a kernel for the dense reward component. 
For throwing, we only take the x-y positions of objects into account during the reward computation, such that our achieved block state is 2-dimensional with $s^i=[s^i_x, s^i_y]\in \mathbb{R}^D$. We evaluate the distance between each block position $s^i$ and the center location $g^i=[g^i_x, g^i_y]$ of each block's goal site individually across the x-y dimensions, such that the sparse component of the reward uses the evaluation:
\begin{equation}
    R_{\text{sparse}}(s^i, g^i) = -1 + \prod_{d=1}^{D}\llbracket|s^i_d-g^i_d|<\delta \rrbracket,
\end{equation}
where $\delta$ is 0.1, corresponding to the half-size of the goal site. The same value is used for $\delta_\text{eval}$ such that we don't require the whole block to be inside the goal site, but the block's center has to be inside the goal site for a successful throw. Note that we do not take the z-position of the object into account for the reward function. The dense reward component is given by:
\begin{equation}
    R_{\text{dense}}(s^i, g^i) = \sum_{d=1}^{D}-1 + \exp\left(-0.5 \cdot |s^i_d-g^i_d| \right).
\end{equation}
For the overall throwing reward, we get the following function with $\eta=0.001$:
\begin{equation}
    R_\text{throw} = \sum_{i=1}^N R_{\text{sparse}}(s^i, g^i) + \eta \cdot R_{\text{dense}}(s^i, g^i).
\end{equation}

\paragraph{\FPP-Flipping}
The flipping task is defined as rotating the blocks $+90^\circ$ around their x-axis. As in the original environment \cite{li2020towards}, the orientation information for each block is encoded in Euler angles in the state vector, where the \texttt{xyz} convention is used. As a result the angle values encode the relative rotating angles about x, y, and z axes in order, \ie after we rotate about x, then we use the new (rotated) y, and the same for z. The flipping task reward thus only applies a constraint on the first Euler angle $\alpha_x$. We use sparse rewards for the flipping task. We also add a small dense component to the reward to incentivize the end effector to stay close to its initialization position $s^\text{init}$. We observed that this additional reward helps the robot find plans for flipping in-place as opposed to flicking objects from the side.

\begin{equation}
    R_\text{flip} = \sum_{i=1}^N \left( -1 + \llbracket \text{dist}(\alpha_x, 90^\circ) < \delta \rrbracket \right) - \eta \cdot \text{dist}(s^\text{agent}, s^\text{init}),
\end{equation}
with $\delta=5^\circ$ and $\eta=0.001$.

\subsubsection{Planning Details for Downstream Tasks}\label{app:planning_extrinsic}
\begin{table}
    \centering
  \caption{Settings for the iCEM controller used for zero-shot generalization in the extrinsic phase of \ourMethod. Same settings were also used for the baseline \mlpicem. Any settings not specified here are the same as the general settings given in \tab{tab:controller_settings}.}
    \label{tab:extrinsic_controller}
    \renewcommand{\arraystretch}{1.03}
    \resizebox{1.\linewidth}{!}{ %
    \begin{tabular}{@{}l|ccccc@{}}
    \toprule
    \textbf{Task}     & \multicolumn{5}{c}{\textbf{Controller Parameters}} \\
                      & Horizon & Colored-noise exponent  & \texttt{use\_mean\_actions} & Noise strength & Cost Along\\
                      & $h$ & $\beta$ &  & $\sigma_\text{init}$ & Trajectory \\
    \midrule
    \Playground-Pushing &   40  &  3.5  &  Yes  &  0.8 &    \texttt{sum}     \\
    \FPP-Stacking      &    30  &  3.5  &   No  &  0.5 &    \texttt{best}     \\
    \FPP-Pick \& Place &    30  &  3.5  &  Yes  &  0.5 &    \texttt{best}     \\
    \FPP-Throwing      &    35  &  2.0  &  Yes  &  0.5 &    \texttt{sum}     \\
    \FPP-Flipping      &    30  &  3.5  &   No  &  0.5 &    \texttt{sum}     \\
  \bottomrule
    \end{tabular}
   }

\end{table}

We use slightly different controller settings for the different tasks as shown in \tab{tab:extrinsic_controller}. These parameters are shared between \ourMethod and the unstructured baseline \mlpicem. 

\subsubsection{Evaluation of Downstream Task Performance}\label{app:performance_extrinsic}

In the \Playground environment, we evaluate the success rate of \ourMethod and the unstructured baseline \mlpicem on the \Playground-Pushing task, when models taken from different training checkpoints are used for planning. Complementary to the results shown in \tab{tab:success_playground} in the main text, \fig{fig:playground:success_temporal} depicts the sample-efficiency of \ourMethod compared to the unstructured baseline \mlpicem. In \fig{fig:playground:task2}, we see that the learned GNN models' ability to capture object-object interactions leads to the selection of more efficient control plans like pushing two objects to the goal position at the same. 

\setcounter{figure}{-1}
\begin{figure}[b!]
    \centering
    \begin{minipage}[c]{.68\linewidth}
    \begin{subfigure}[t]{.32\linewidth}
    \includegraphics[width=\linewidth]{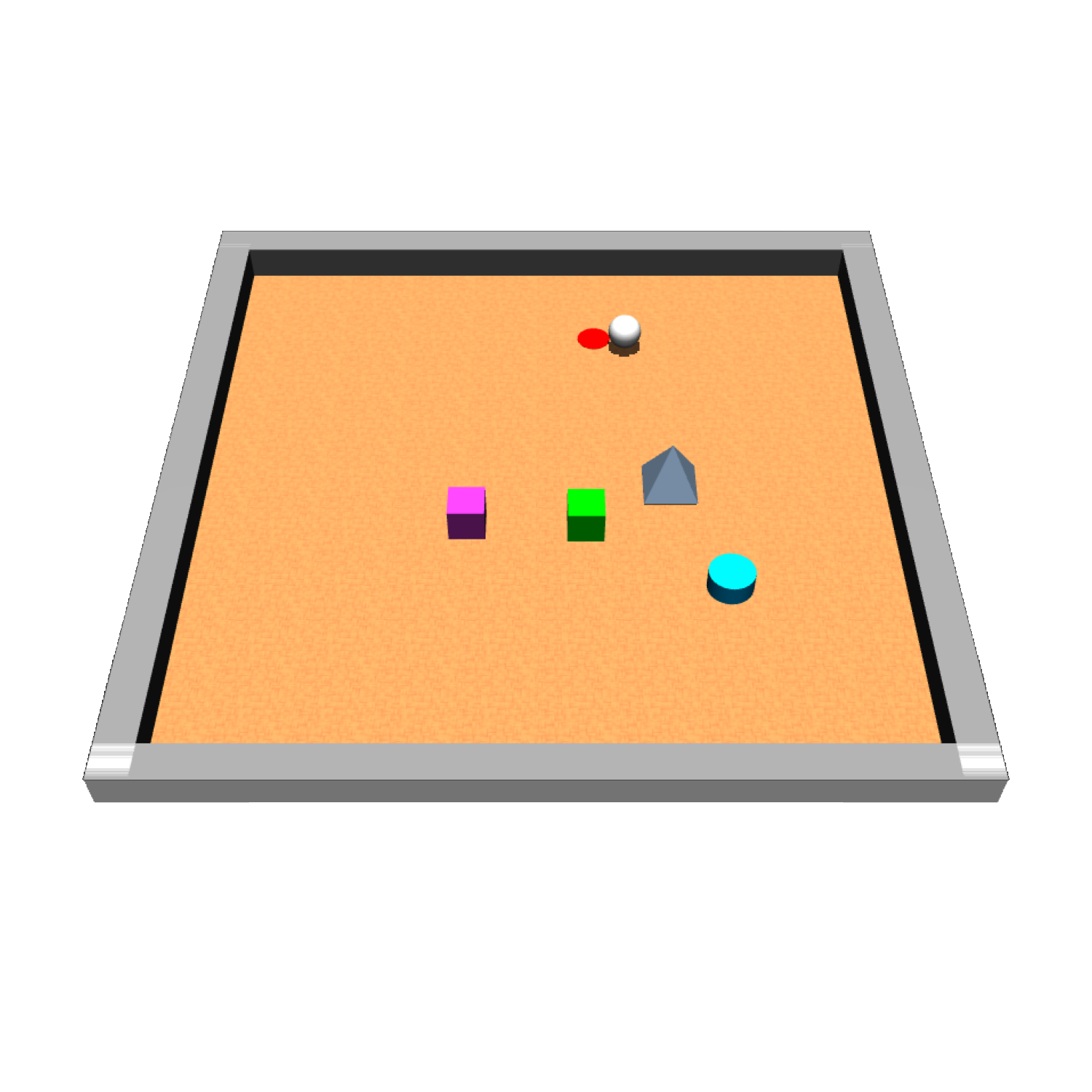}
    \vspace{-3.em}  
    \caption{$t=0$}
    \end{subfigure}
    \begin{subfigure}[t]{.32\linewidth}
    \includegraphics[width=\linewidth]{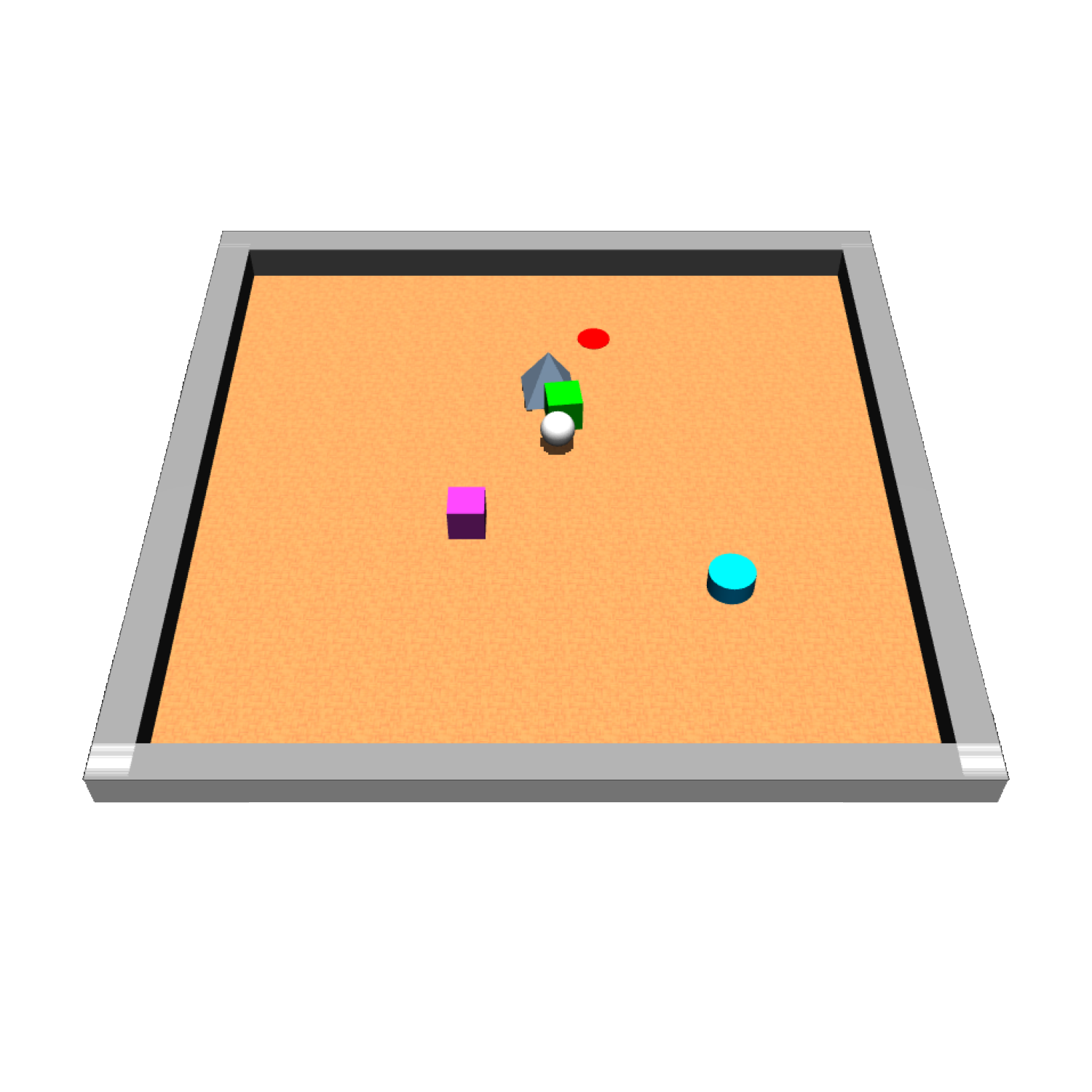}
    \vspace{-3.em}  
    \caption{$t=100$}
    \end{subfigure}
    \begin{subfigure}[t]{.32\linewidth}
    \includegraphics[width=\linewidth]{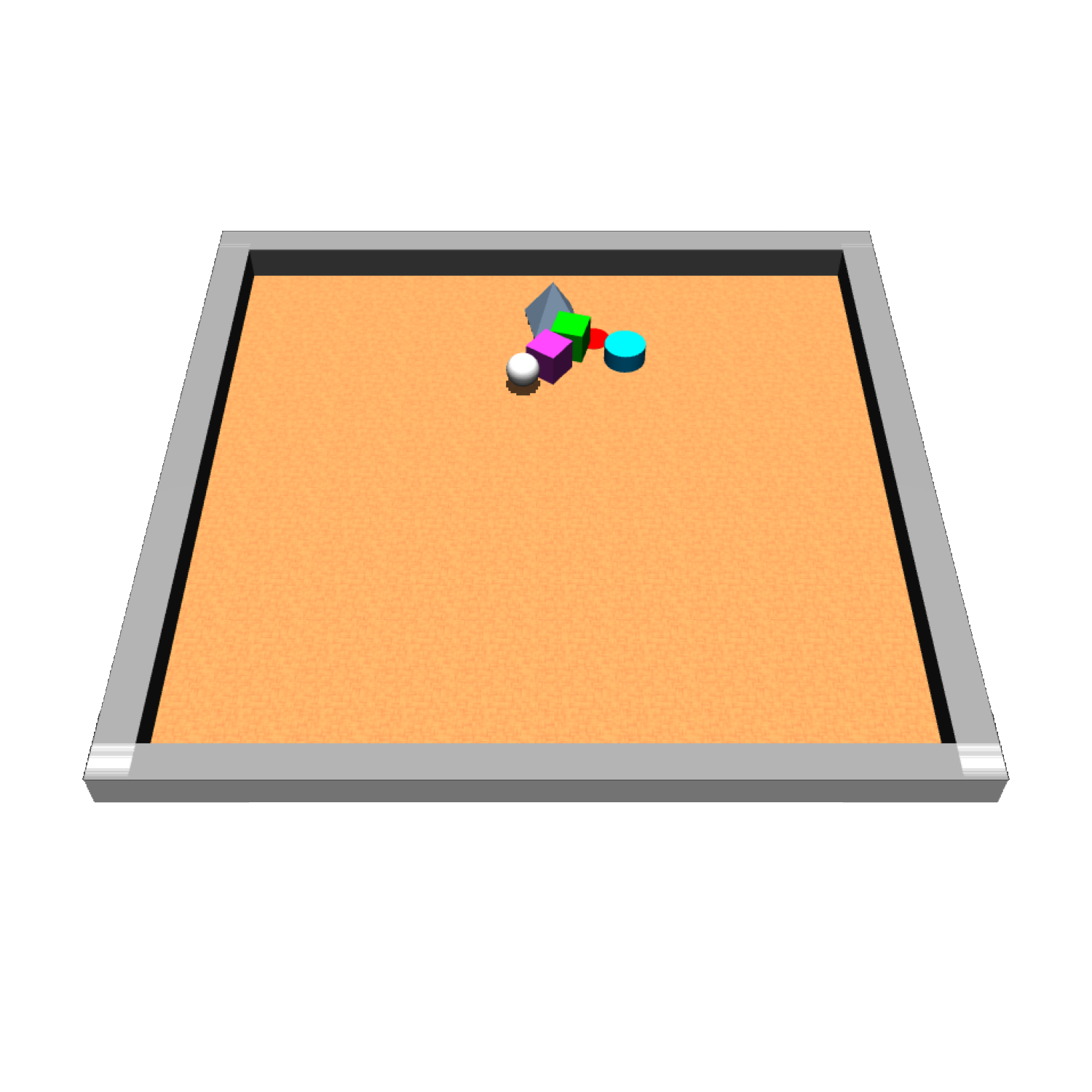}
    \vspace{-3.em}  
    \caption{$t=300$}
    \end{subfigure}\vspace{-.5em}
    \captionof{figure}{Downstream task \Playground-Pushing: move all objects to target location (red), as solved by \ourMethod. At $t=100$, we see that the planner can find a trajectory pushing two objects at the same time with the learned GNN models. At $t=300$, the task is already solved.}
    \label{fig:playground:task2}
    \end{minipage}
    \hfill
    \setcounter{figure}{0}
    \begin{minipage}[c]{.30\linewidth}
    \begin{subfigure}[t]{.98\linewidth}
    \includegraphics[width=\linewidth]{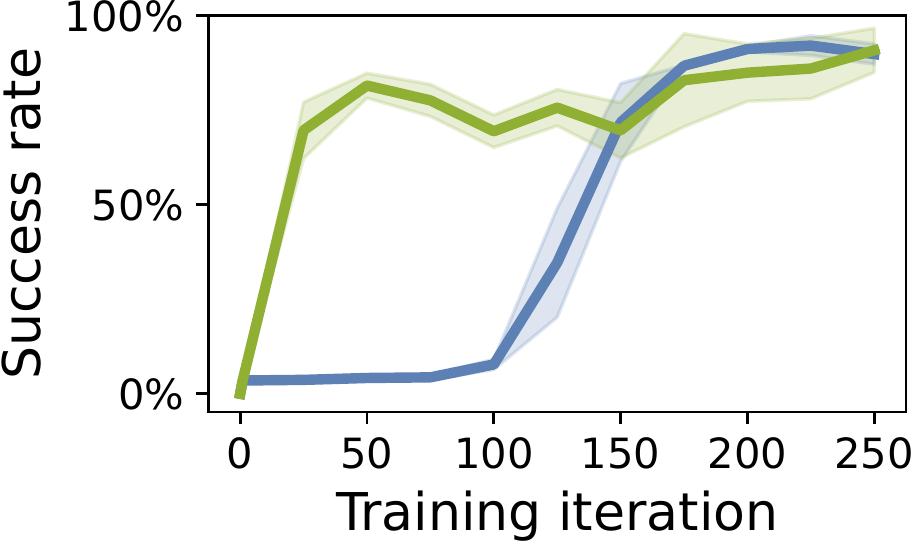}
    \end{subfigure}
    \captionof{figure}{Success rate of downstream task in \Playground for {\color{ourgreen}\ourMethod}  vs. {\color{ourblue}\mlpicem} evaluated at different training iterations.}
    \label{fig:playground:success_temporal}
    \end{minipage}
\end{figure}

\tab{tab:fpp:performance_stack} and \tab{tab:fpp:performance_throw_flip} contain the success rates reported for zero-shot generalization on downstream tasks in the \FPP environment, complementary to \fig{fig:fpp:performance} in the main text.

\setlength{\intextsep}{1.0pt plus 2.0pt minus 2.0pt}
\setlength{\columnsep}{10pt}%
\begin{wrapfigure}{r}{0.3\textwidth}
    \centering
    \includegraphics[width=\linewidth]{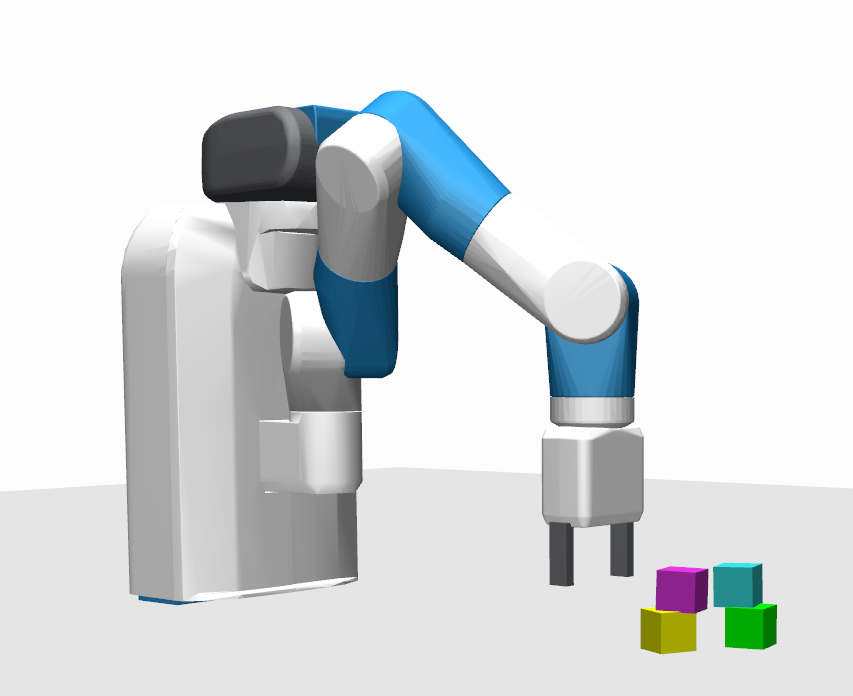}\vspace{-.1em}
    \caption{Multi-tower stacking task (Here task solved by \ourMethod).}
    \label{fig:fpp:multitower}
\end{wrapfigure}

We spawn the environment with the number of objects specified in the table. The \mlpicem baseline that lacks combinatorial generalization, can only be applied to the 4 object case, as seen during the free-play phase. We do not perform masking of objects during planning for this baseline, as we consider a task with \eg 2 object stacking to be defined in an environment spawned with the same amount of objects.
For \FPP-Stacking, in addition to the single-tower stacking, we also test for the multi-tower task with 4 objects (denoted by $2+2$) such that the goal is to build two towers with 2 blocks each. Since the base of these towers can be close to one another, this task has an increased level of difficulty compared to stacking 2 blocks, also reflected in the success rates shown in \tab{tab:fpp:performance_stack}.

In the stacking tasks (single-tower and multi-tower), success is 1, only when the required towers with all the objects present in the environment are fully stacked. In all the other tasks in \FPP and in \Playground-Pushing, the success rate in the multi-object setup is defined as the fraction of objects solved relative to the total number of objects spawned in the environment. For example, in an environment with 4 objects, $0.75$ success rate means 3 out of 4 objects reached their respective goal positions.

\begin{table}
    \centering
    \caption{Zero-shot generalization performance of \ourMethod vs. \mlpicem on downstream tasks \FPP-Pick\&Place and \FPP-Stacking.}
    \label{tab:fpp:performance_stack}
    \renewcommand{\arraystretch}{1.03}
    \resizebox{1.\linewidth}{!}{ %
    \begin{tabular}{@{}l|ccccc|cccc@{}}
    \toprule
    \textbf{Task}   & \multicolumn{5}{c}{Pick\&Place} & \multicolumn{4}{c}{Stacking}\\
    \#~Objects    & 2 & 3 & 4 & 5 & 6 & 2 & 3 & 4 & 2+2 \\
    \midrule
    \ourMethod & $0.94 \pm 0.01$  & $0.96 \pm 0.01$  & $0.97 \pm 0.01$ & $0.96 \pm 0.01$ & $0.96 \pm 0.001$ &
     $0.91 \pm 0.02$& $0.27 \pm 0.08$ & $0.02 \pm 0.02$ & $0.236 \pm 0.06$ \\
     MLP-iCEM   & - & - & $0.83 \pm 0.02$  & -  & - 
     & - & - & $0 \pm 0$ & $0.013 \pm 0.012$ \\
  \bottomrule
    \end{tabular}
    }
  \vspace{.3cm}
    \centering
    \caption{Zero-shot generalization performance of \ourMethod vs. \mlpicem on downstream tasks \FPP-Throwing and \FPP-Flipping.}
    \label{tab:fpp:performance_throw_flip}
    \renewcommand{\arraystretch}{1.03}
    \resizebox{1.\linewidth}{!}{ %
    \begin{tabular}{@{}l|ccc|ccccc@{}}
    \toprule
    Task & \multicolumn{3}{c}{Throwing} & \multicolumn{5}{c}{Flipping} \\
    \#~Objects    & 2 & 3 & 4 & 2 & 3 & 4 & 5 & 6\\
    \midrule
    \ourMethod & $0.675 \pm 0.039$ & $0.67 \pm 0.04$ & $0.61 \pm 0.01$ &
    $0.94 \pm 0.01$ & $0.93 \pm 0.03$ & $0.88 \pm 0.04$  & $0.87 \pm 0.04$ & $0.83 \pm 0.04 $\\
     MLP-iCEM   & - & - & $0.32 \pm 0.03$ & - & - & $0.25 \pm 0.02$ & - & - \\
  \bottomrule
    \end{tabular}
    }
\end{table}

\subsection{Hyperparameter settings for baselines}\label{app:baselines}
The hyperparameters for the model architecture and the training of the unstructured baseline \mlpicem, which corresponds to \ourMethod without GNNs, are given in \tab{tab:mlp_model_training_settings}.
\begin{table*}[htb]
    \centering
    \caption{Base settings for MLP model training in \mlpicem.}
    \label{tab:mlp_model_training_settings}
    \begin{subtable}[t]{.5\textwidth}
        \centering
        \caption{General settings.}
        \begin{tabular}{@{}ll@{}}
        \toprule
        \textbf{Parameter} & \textbf{Value} \\
        \midrule
        Network Size & $3 \times 256$ \\
        Activation function & SiLU \\
        Ensemble Size & 5\\
        Optimizer & ADAM \\
        Batch Size & $256$ \\
        Epochs & $50$ \\
        Learning Rate & $0.0001$ \\
        Weight decay & $5 \cdot 10^{-5}$ \\
        Weight Initialization & Truncated Normal\\
        Normalize Input & Yes \\
        Normalize Output & Yes \\
        Predict Delta & Yes \\
        \bottomrule
        \end{tabular}
    \end{subtable}%
    \begin{subtable}[t]{.5\textwidth}
        \centering
        \caption{Environment-specific settings.}
        \begin{tabular}{@{}ll@{}}
        \toprule
        \multicolumn{2}{c}{\Playground} \\
        \textbf{Parameter} & \textbf{Value} \\
        \midrule
        Network Size & $3 \times 128$ \\
        Batch Size & $128$ \\
        \bottomrule
        \\
        \toprule
        \multicolumn{2}{c}{\FPP} \\
        \textbf{Parameter} & \textbf{Value} \\
        \midrule
        \multicolumn{2}{c}{Same as general settings} \\
        \bottomrule
        \end{tabular}
    \end{subtable}
\end{table*}

For the other baselines RND~\cite{burda2018exploration}, Disagreement~\cite{pathak2019self} and ICM~\cite{pathak2017curiosity}, we use the implementation from \citet{laskin2021urlb} that uses DDPG \cite{Lillicrap2016DDPG} with the same hyperparameter settings proposed there. The code for these baselines can be found in \url{https://github.com/rll-research/url_benchmark}.

\subsection{Offline RL}\label{app:offline_rl}

\begin{table*}[htb]
    \centering
    \caption{Settings for offline RL.}
    \label{tab:rl_ddpg_hyperparams}
    \begin{subtable}[t]{.5\textwidth}
        \centering
        \caption{Settings for CQL.}
        \begin{tabular}{@{}ll@{}}
        \toprule
        \textbf{Parameter} & \textbf{Value} \\
        \midrule
        Batch size & 256\\
        Actor learning rate & 1.0e-4\\
        Critic learning rate & 3.0e-4\\
        Temp learning rate & 1.0e-4\\
        Alpha learning rate & 0.0\\
        Conservative weight & 10.0\\
        Number of action samples & 10\\
        \texttt{q\_func\_factory} & mean\\
        Optimizer & ADAM \\
    Actor Encoder Network Size & $3 \times 256$ \\
    Critic Encoder Network Size & $3 \times 256$ \\
        \texttt{gamma} & 0.99\\
        \texttt{tau} & 0.005\\
        \texttt{n\_critics} & 2\\
        Initial temperature & 1.0\\
        Initial $\alpha$ & 1.0\\
        $\alpha_\text{threshold}$ & 10.0\\
        Conservative weight & 1.0\\
        \texttt{soft\_q\_backup} & No\\
        \bottomrule
        \end{tabular}
    \end{subtable}%
    \begin{subtable}[t]{.5\textwidth}
        \centering
        \caption{HER Settings}
        \begin{tabular}{@{}ll@{}}
        \toprule
        \multicolumn{2}{c}{\Playground} \\
        \textbf{Parameter} & \textbf{Value} \\
        \midrule
        Replay Strategy & Future \\
        \texttt{replay\_k} & $4$ \\
        \bottomrule
        \end{tabular}
    \end{subtable}
\end{table*}

\paragraph{Policy Selection} Since CQL tends to overfit to the training data, resulting in a significant drop in task performance, we use early stopping to select the best policy on a run by run basis.

\paragraph{Rewards} To train the policies with offline RL, we use sparse rewards in all the experiments. The reward is computed according to:
\begin{equation}
    r(g^{\textrm{achieved}}, g^{\textrm{desired}}) = \llbracket \textrm{dist}(g^{\textrm{achieved}}, g^{\textrm{desired}}) < \delta \rrbracket \in [0, 1],
\end{equation}
with \( g^{\textrm{achieved}} \) being the achieved goal, \( g^{\textrm{desired}} \) being the desired goal and \( \delta \) being a task-dependent threshold. Depending on the task, \( g^{\textrm{achieved}} \) is equal to \( s^{\textrm{agent}} \) or \( s^{i} \), \( i=1,\ldots,N \), or any combination of these.

\paragraph{State Representation} In \Playground{}, we use the flat state representation provided by the environment as input for the CQL algorithm. In \FPP, we also use the flat state representation provided by the environment, including the relative positions of the objects to the end-effector position. Without this relative information, CQL could not learn a policy for the object manipulation task.

\paragraph{Datasets} We use the data collected by the different intrinsically motivated agents during free play as datasets for offline RL. The same amount of free-play data is used from the different agents to generate the datasets. In \Playground{}, the datasets contain \( 500,000 \) transitions. In \FPP{}, the datasets contain \( 600,000 \) transitions.

\paragraph{Tasks} In \Playground{}, we evaluate the performance of offline RL trained with the different datasets on two tasks: (i) move the agent to a randomly sampled target location and (ii) move the first object to a randomly sampled target location. In \FPP{}, we evaluate CQL on (i) move the end-effector to a randomly sampled location and (ii) move the first object to a randomly sampled target location that is in the air in \( 50\% \) of the cases.

\subsection{Uncertainty Heatmaps}\label{app:uncertainty_heatmap}

\begin{figure}
\center
\includegraphics[width=.95\textwidth]{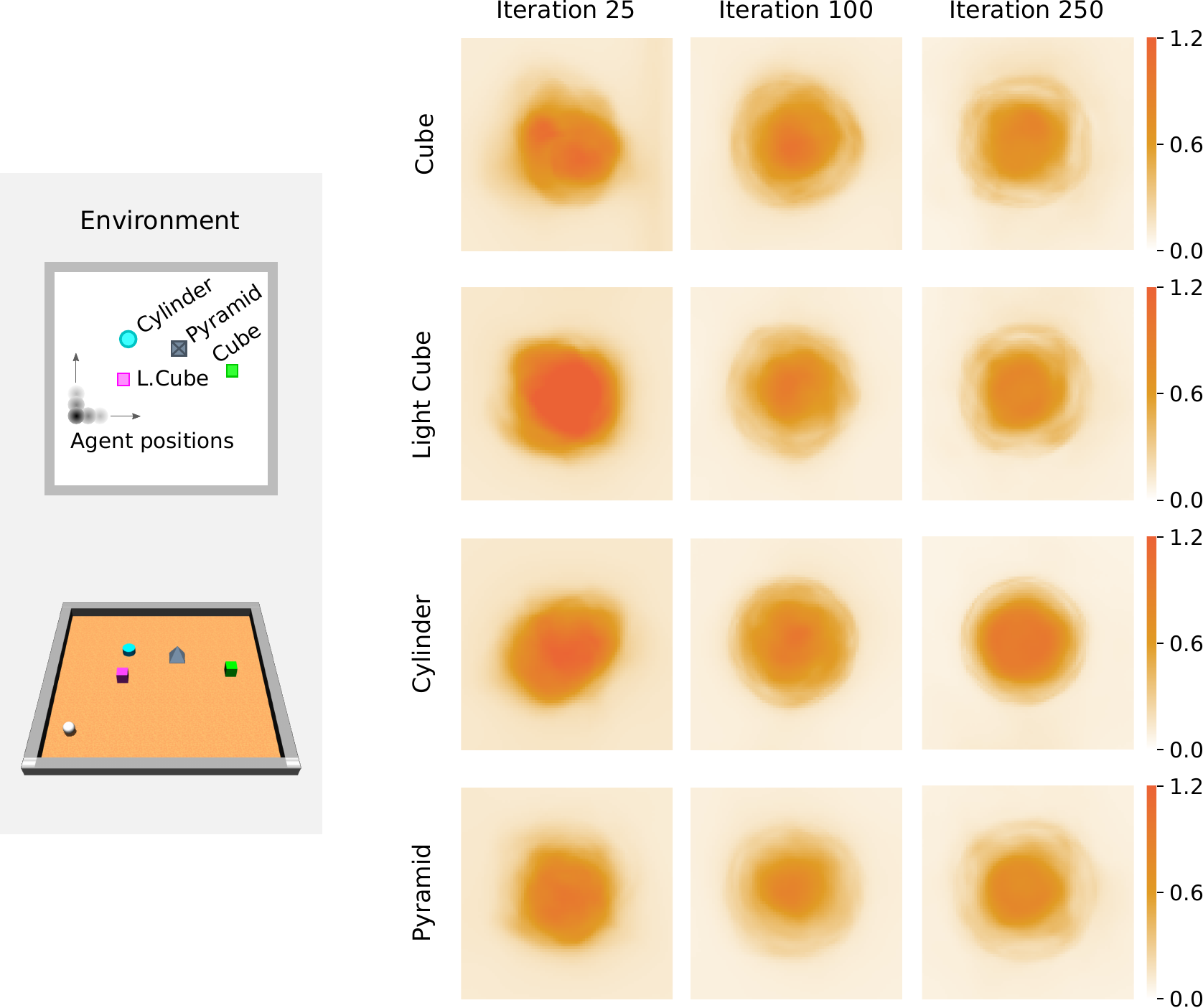}\vspace{-.1em}
\caption{\textbf{Close-up of each object in uncertainty heatmaps of the GNN ensemble with \ourMethod in \Playground{}.} }
\label{fig:object_centric_uncertainty}
\end{figure}

The uncertainty heatmaps are computed by a spatial discretization of the playground area (bin size equal to $r/10$, where $r$ is the agent's radius) and evaluating \eqn{eqn:epistemic} for 8 actions, equidistant unit vectors on the unit circle, for the agent hypothetically being at each $(x,y)$ location of the grid. This means that we also spawn the agent inside objects and walls, violating the regular physical properties of the environment. As a result, there is always remaining uncertainty inside the walls and inside the objects.

The progression of the uncertainty heatmaps shown in \fig{fig:gnn_vs_mlp} can be interpreted as follows:
\begin{itemize}
    \item \textbf{Iteration 1}: After only one training iteration of \ourMethod, we still have uniform uncertainty of the model as the model lacks training and the model's predictions are very inaccurate.
    \item \textbf{Iteration 25}: As more data of agent-object interactions are collected, the model discovers objects as a source of uncertainty, resulting in high epistemic uncertainty around the objects. As the models also start learning that the agent cannot permeate objects, \ie agent and object cannot occupy the same space, the uncertainty at the center of objects is also high. Notice how this isn't the case for the variant \mlpicem (see \fig{fig:gnn_vs_mlp}). As the MLP models lack data from agent-object interactions, the model is confident (low ensemble disagreement) that the agent can simply move through objects. Same principle also applies to the walls. Before \ourMethod generates enough agent-wall interactions, there is no reason for the agent to expect any different dynamics at the boundaries of the playground since the walls are not part of the state information. For instance, the GNN ensemble doesn't have enough data generated at the left wall at iteration 25, resulting in low ensemble disagreement.
    \item \textbf{Iterations 100~-~249}: As \ourMethod generates more agent-object and object-object interactions, the uncertainty around the object boundaries starts decreasing as the GNN ensemble learns more about each object's dynamics. As demonstrated in \fig{fig:object_centric_uncertainty}, the objects' shapes (cube, cylinder, pyramid) becomes discernible with more training iterations. As explained above, the uncertainty at the center of each object always remains.
\end{itemize}

\begin{figure}
\center
\includegraphics[width=.95\textwidth]{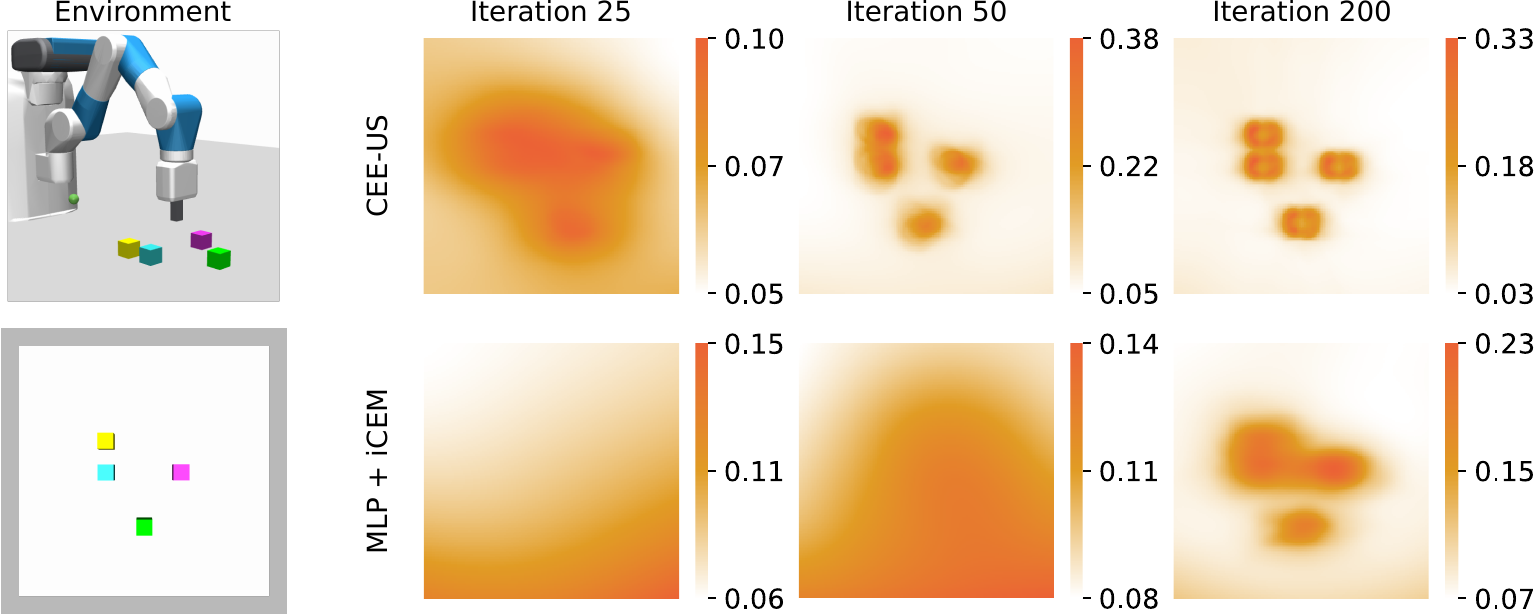}\vspace{-.1em}
\caption{\textbf{Uncertainty heatmaps of the world model ensembles with \ourMethod in \FPP{}, compared to the unstructured baseline \mlpicem.} Similar to the results shown in the \Playground environment, the uncertainty in \FPP is also localized around the objects early on in the training, with object shapes becoming even more discernible further along in free play at iteration 200.}
\label{fig:fpp_uncertainty}
\end{figure}

We generate similar uncertainty heatmaps for the \FPP environment as shown in \fig{fig:fpp_uncertainty}. Here, we only put the robot gripper on different locations on the table, more precisely on a 80 cm $\times$ 80 cm square grid around the initial gripper position, that is discretized into bins of size 0.005 mm corresponding to $1/10$th of the cube size. In order to obtain the uncertainty heatmaps, we evaluate the ensemble disagreement \eqn{eqn:epistemic} for 100 random actions at each hypothetical location of the robot arm in the spatially discretized grid on the table.

\section{Multi-step Prediction Performance of GNNs and MLPs}\label{app:gnn_pred}
\begin{figure}[!b]
\center
\includegraphics[width=.95\textwidth]{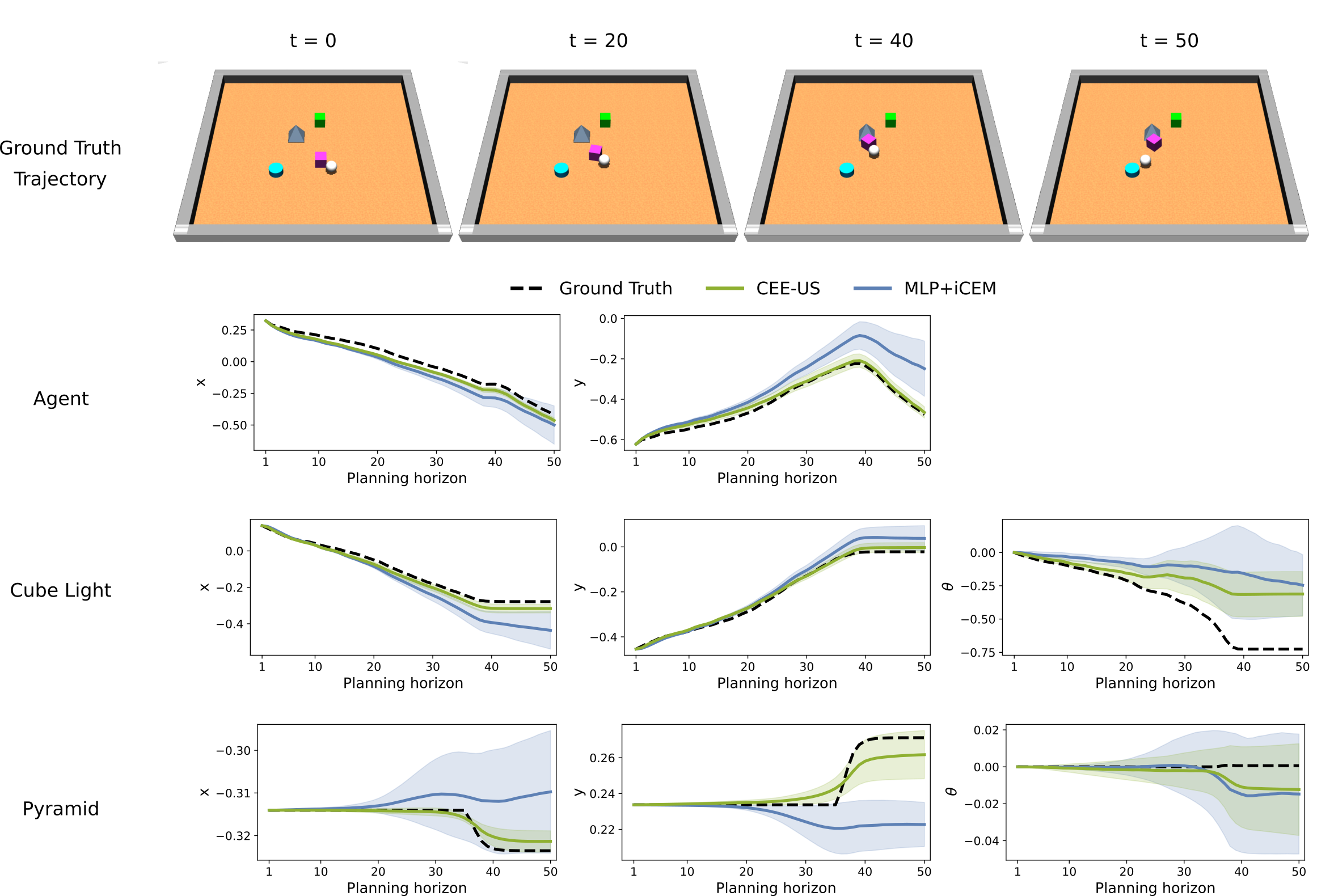}\vspace{-.1em}
\caption{\textbf{Multi-step prediction performance of the trained GNN ensemble with \ourMethod at the end of free play compared to the trained MLP ensemble with \mlpicem.}}
\label{fig:multi_step_prediction}
\end{figure}

In the \Playground environment, we showcase the multi-step prediction performance of the trained GNN vs. MLP dynamics models at the end of free play. For a given starting state of the environment at $t=0$ and an action sequence $(a_t)_{t=0}^{H-1}$, we generate a rollout in imagination of the trained models and compare these multi-step dynamics predictions to the ground truth. An example trajectory can be found in \fig{fig:multi_step_prediction}. We compute the cumulative prediction error of the generated trajectories, taking the mean predictions across the ensemble members for the GNNs and MLPs respectively, on 50 random evaluation rollouts with a multi-step prediction horizon of 50 timesteps. The evaluation rollouts are generated using a random policy, that interacts with one or more randomly chosen objects at each rollout. Using the GNN ensemble we get a prediction error of 2.82, whereas for the MLP ensemble we get 4.06 (cumulative error over the time horizon as well as the state space dimension). This illustrates the improved dynamics prediction that is obtained through the use of structured world models.

\begin{figure}
    \centering
    {\scriptsize
      \textcolor{ours}{\rule[2pt]{20pt}{1pt}} \ourMethod{} \quad \textcolor{baseline1}{\rule[2pt]{20pt}{1pt}} \mlpicem 
    }\\
    \begin{subfigure}[b]{0.01\textwidth}
        \notsotiny{\rotatebox{90}{\hspace{.5cm}\textsf{relative time}}}
    \end{subfigure} 
    \begin{subfigure}[t]{.24\linewidth}
    \includegraphics[width=\linewidth]{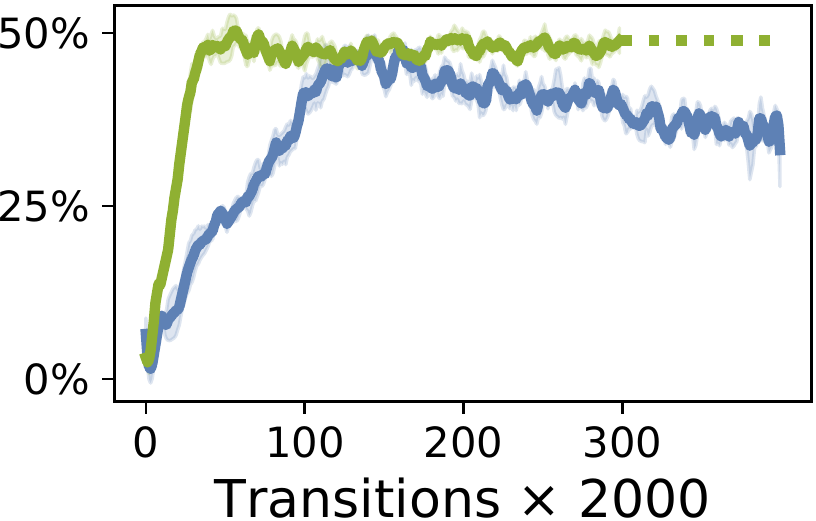}\vspace{-.3em}
    \caption{1 object moves}
    \end{subfigure}
    \hfill
    \begin{subfigure}[t]{.24\linewidth}
    \includegraphics[width=\linewidth]{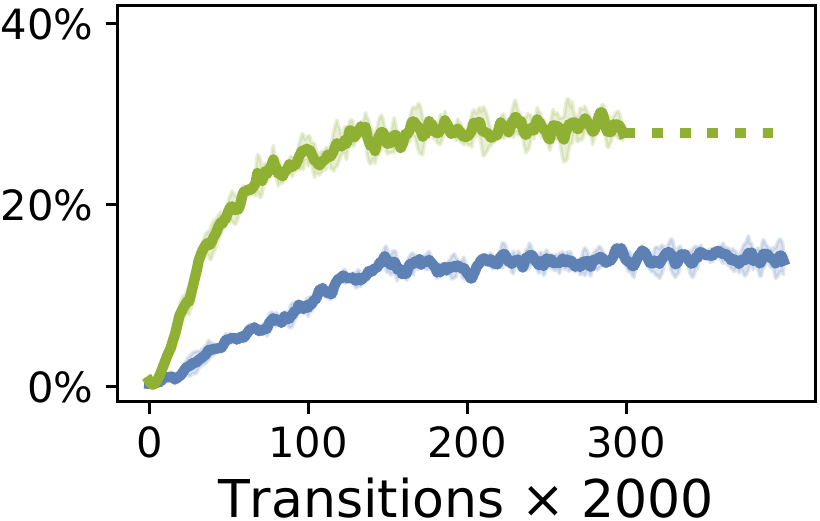}\vspace{-.3em}
    \caption{2 or more objects move}
    \end{subfigure}\hfill
    \begin{subfigure}[t]{.24\linewidth}
    \includegraphics[width=\linewidth]{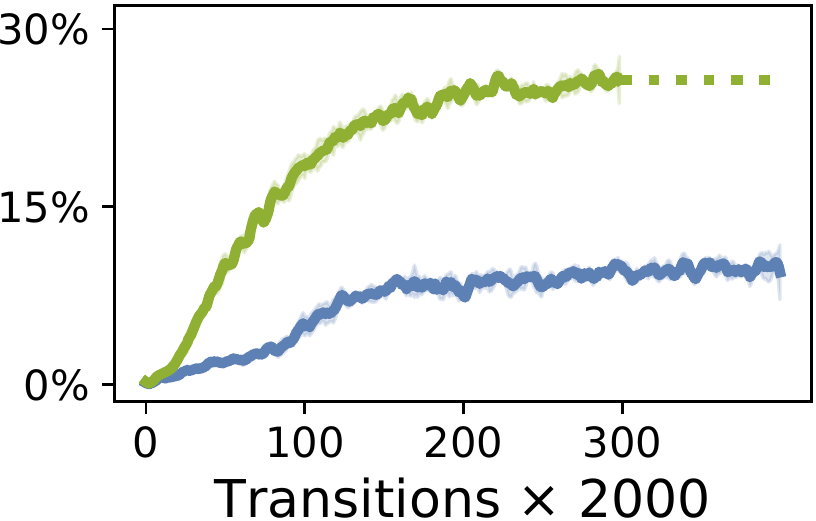}\vspace{-.3em}
    \caption{object(s) flipped}
    \end{subfigure}
    \begin{subfigure}[t]{.24\linewidth}
    \includegraphics[width=\linewidth]{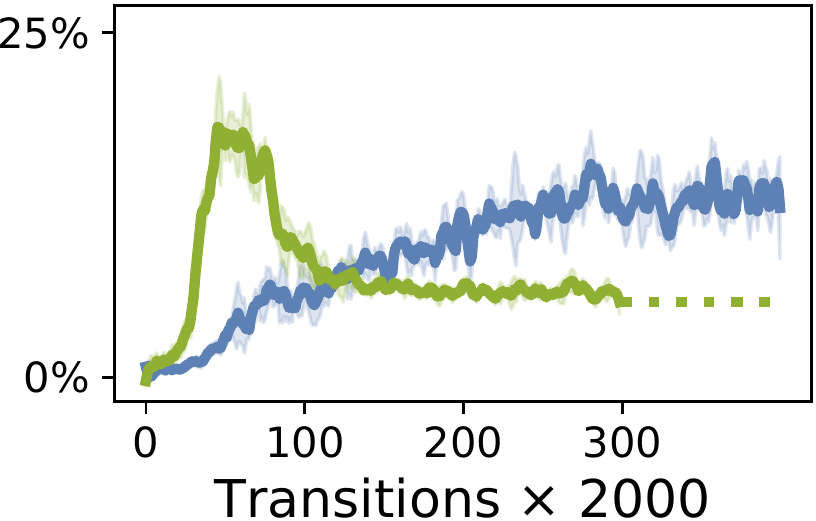}\vspace{-.3em}
    \caption{object(s) in air}
    \end{subfigure}
    \caption{{\bf Asymptotic behavior of \mlpicem during free play in \FPP compared to \ourMethod{}}.
     Interaction metrics of free-play exploration count the relative amount of time steps spent in moving one object (a), moving two and more objects (b), flipping object(s) (c), and moving objects into the air (d). This is the extension of \fig{fig:interactions:FPP} in the main paper. Here the baseline \mlpicem is trained for an additional 100 training iterations (corresponding to 200K more transitions collected). For \ourMethod, we use the dashed line to visualize the achieved interaction metrics at the end of training at iteration 300. Three independent seeds were used.
    }
    \label{fig:interactions_continued:FPP}
\end{figure}

\begin{figure}
    \centering
    {\scriptsize
      \textcolor{ours}{\rule[2pt]{20pt}{1pt}} \ourMethod{} \quad \textcolor{baseline1}{\rule[2pt]{20pt}{1pt}} \mlpicem 
    }\\
    \begin{subfigure}[t]{.33\linewidth}
    \includegraphics[width=\linewidth]{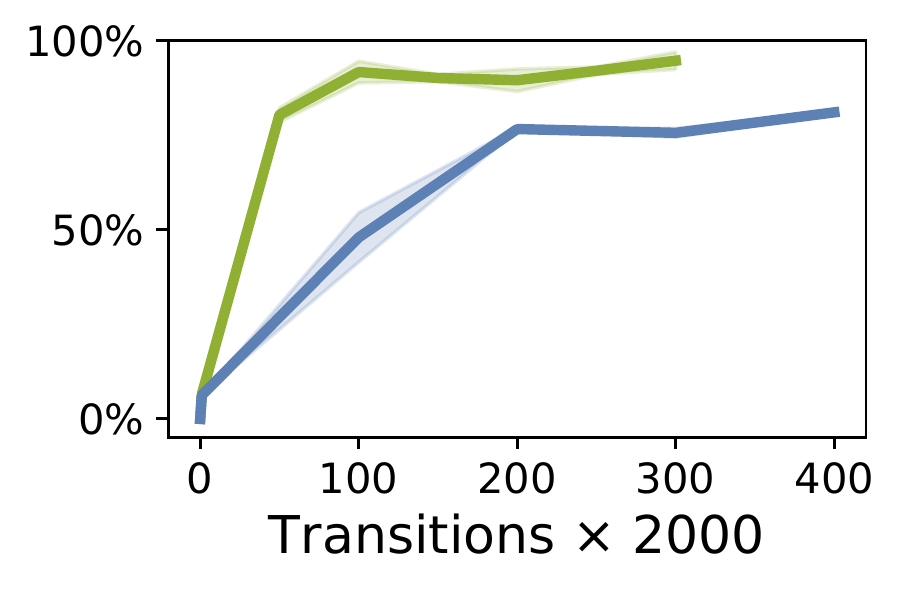}\vspace{-.3em}
    \caption{Pick \& Place 4 Objects}
    \end{subfigure}
    \begin{subfigure}[t]{.33\linewidth}
    \includegraphics[width=\linewidth]{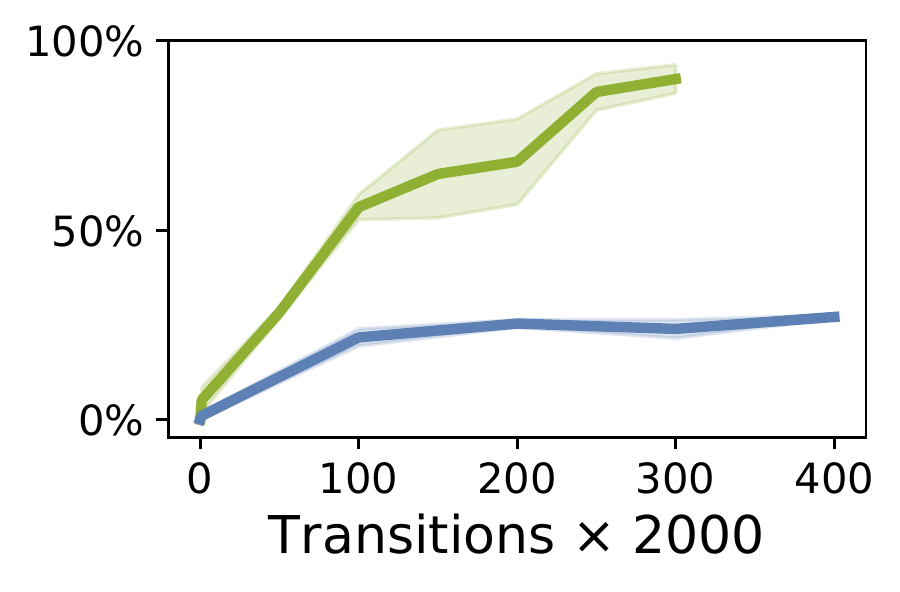}\vspace{-.3em}
    \caption{Flip 4 Objects}
    \end{subfigure}
    \caption{{\bf Asymptotic downstream task performance of \mlpicem in \FPP compared to \ourMethod}.
     We compute the achieved success rates for models checkpointed over the course of free play, where we train the \mlpicem for an additional 100 training iterations (corresponding to 200K more data points collected). Three independent seeds were used.
    }
    \label{fig:task_continued:FPP}
\end{figure}

In \fig{fig:interactions_continued:FPP}, we also show the behavior of \mlpicem later on during free play in terms of interaction metrics, where we train it for an additional 100 iterations. In \fig{fig:task_continued:FPP}, we show the downstream task performance on the Pick \& Place and Flipping tasks for models checkpointed at different iterations of free play. Even with an additional 100 iterations of free play, \mlpicem's downstream task performance is inferior to \ourMethod. Overall, this showcases the importance of the accurate forward dynamics prediction of GNNs not just in terms of sample-efficiency of interaction metrics, but also for zero-shot downstream task generalization. 

\section{Combining Model-based Control with Random Network Distillation} \label{app:model_rnd}
One key element of our method is that we use the ensemble disagreement to approximate the epistemic uncertainty of the model itself. This is inherently different from the intrinsic rewards computed for instance in Random Network Distillation (RND) \cite{burda2018exploration}. RND is essentially an expansion of count-based methods to continuous domains and the intrinsic reward is decoupled from the actual model performance of the dynamics model. Note that in the case of RND, as explained in \sec{interaction}, the RND module tries to match the output of a random target network. As long as a state is not visited enough, the RND module will generate high intrinsic reward regardless of whether the model can already predict this state accurately or not. In the opposite scenario, even if a state is trivial, the RND module is agnostic to the invariances and symmetries in an environment. As a result, it will try to create state-space coverage even when the state transition dynamics is trivial to learn. 
In order to test how using the RND intrinsic reward for planning affects behavior during free play, as well as the consequent downstream task performance, we ran new baselines that we refer to as \gnnrndicem and \mlprndicem. In these baselines, we have a world model learning the dynamics (GNN for \gnnrndicem and an MLP for \mlprndicem) which is used for model-based planning during free play. However, instead of using ensemble disagreement, we use the intrinsic reward of a separate RND module (MLP) to structure the free play. The RND network is also trained on the generated free-play data, separately from the actual world model. 

\begin{figure}
    \centering
    {\scriptsize
      \textcolor{ours}{\rule[2pt]{20pt}{1pt}} \ourMethod{} \quad \textcolor{baseline1}{\rule[2pt]{20pt}{1pt}} \mlpicem \quad  \textcolor{baseline3}{\rule[2pt]{20pt}{1pt}} \gnnrndicem
       \quad \textcolor{baseline4}{\rule[2pt]{20pt}{1pt}} \mlprndicem
    }\\
    \begin{subfigure}[b]{0.01\textwidth}
        \notsotiny{\rotatebox{90}{\hspace{.5cm}\textsf{relative time}}}
    \end{subfigure} 
    \begin{subfigure}[t]{.24\linewidth}
    \includegraphics[width=\linewidth]{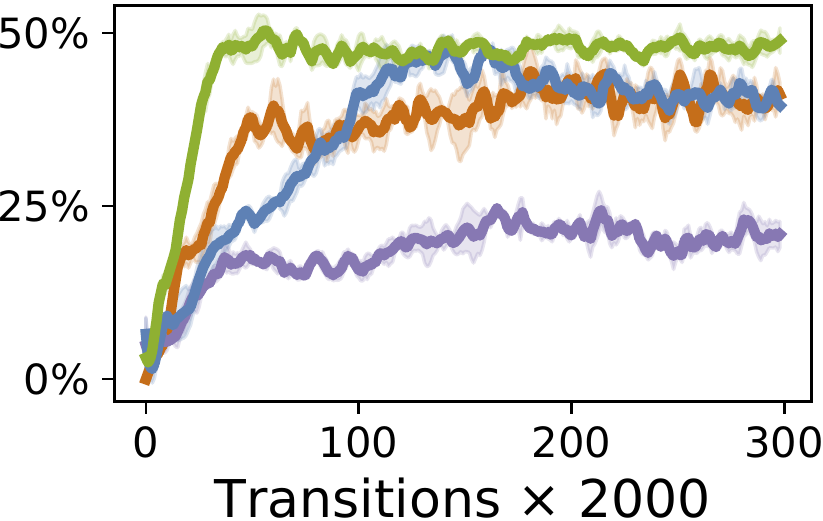}\vspace{-.3em}
    \caption{1 object moves}
    \end{subfigure}
    \hfill
    \begin{subfigure}[t]{.24\linewidth}
    \includegraphics[width=\linewidth]{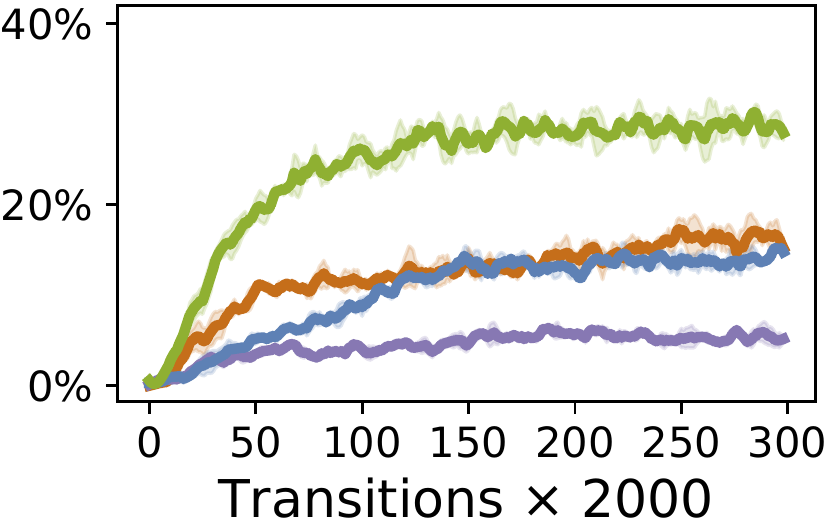}\vspace{-.3em}
    \caption{2 or more objects move}
    \end{subfigure}\hfill
    \begin{subfigure}[t]{.24\linewidth}
    \includegraphics[width=\linewidth]{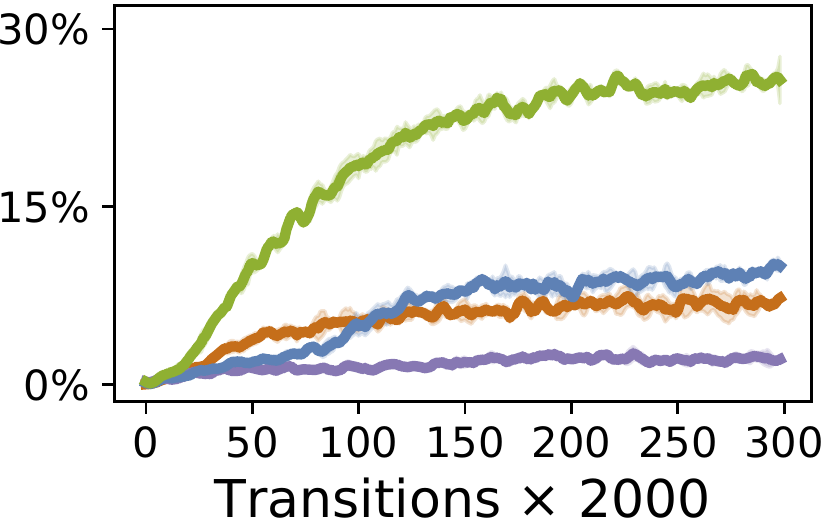}\vspace{-.3em}
    \caption{object(s) flipped}
    \end{subfigure}
    \begin{subfigure}[t]{.24\linewidth}
    \includegraphics[width=\linewidth]{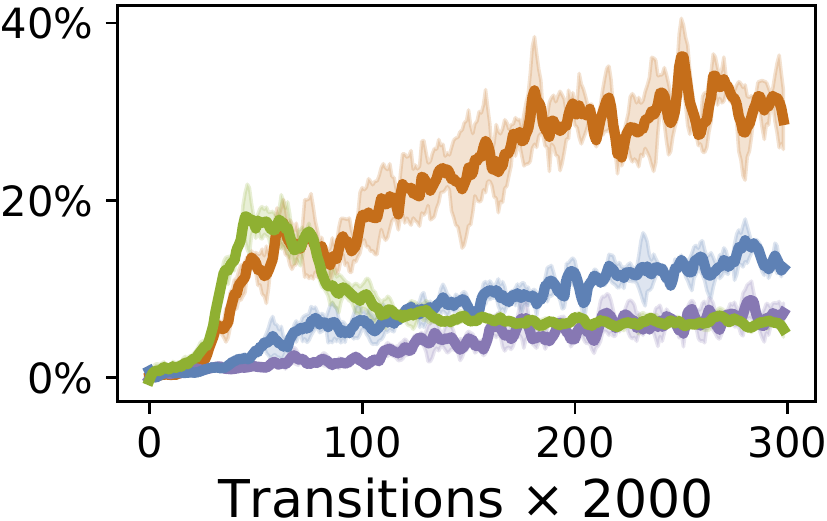}\vspace{-.3em}
    \caption{object(s) in air}
    \end{subfigure}
    \caption{{\bf Interactions generated during free play in \FPP by \ourMethod{}}.
     Interaction metrics of free-play exploration count the relative amount of time steps spent in moving one object (a), moving two and more objects (b), flipping object(s) (c), and moving objects into the air (d). We compare \ourMethod{} and \mlpicem, which use the epistemic uncertainty of the model approximated via the ensemble disagreement, with variants that use a separate RND module for intrinsic reward. We used three independent seeds.
    }
    \label{fig:interactions:FPP_wRND}
\end{figure}

\begin{figure}
    \centering
    {\scriptsize
      \textcolor{ours}{\rule[2pt]{20pt}{1pt}} \ourMethod{} \quad \textcolor{baseline1}{\rule[2pt]{20pt}{1pt}} \mlpicem \quad  \textcolor{baseline3}{\rule[2pt]{20pt}{1pt}} \gnnrndicem
       \quad \textcolor{baseline4}{\rule[2pt]{20pt}{1pt}} \mlprndicem
    }\\
    \begin{subfigure}[b]{0.01\textwidth}
        \notsotiny{\rotatebox{90}{\hspace{.5cm}\textsf{success rate}}}
    \end{subfigure} 
    \begin{subfigure}[t]{.24\linewidth}
    \includegraphics[width=\linewidth]{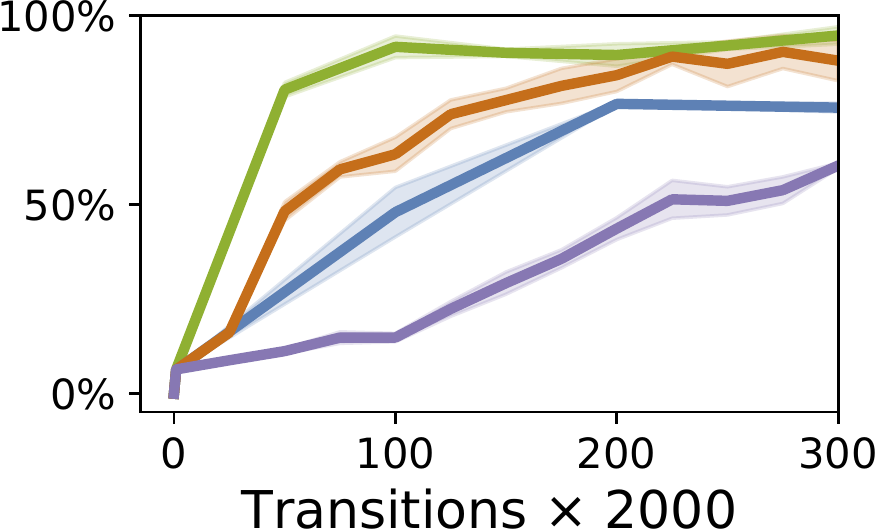}\vspace{-.3em}
    \caption{Pick \& Place 4 Objects}
    \end{subfigure}
    \hfill
    \begin{subfigure}[t]{.24\linewidth}
    \includegraphics[width=\linewidth]{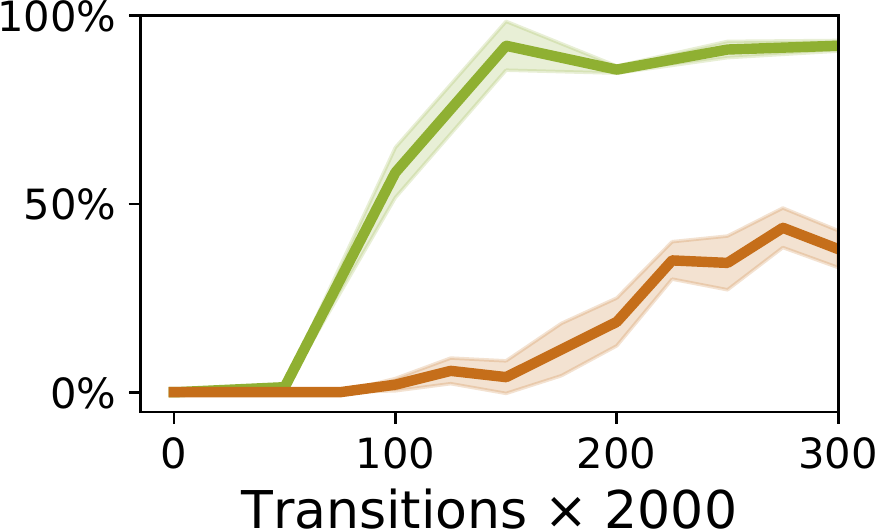}\vspace{-.3em}
    \caption{Stack 2 Objects}
    \end{subfigure}\hfill
    \begin{subfigure}[t]{.24\linewidth}
    \includegraphics[width=\linewidth]{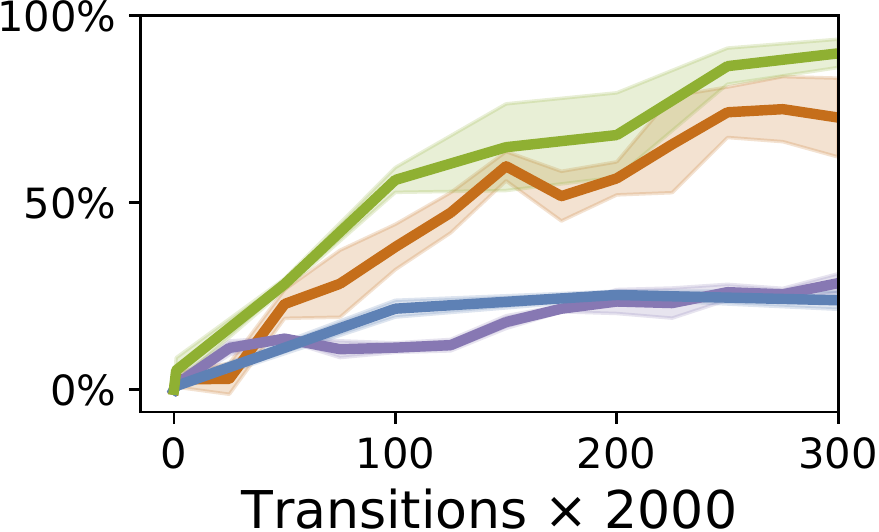}\vspace{-.3em}
    \caption{Flip 4 Objects}
    \end{subfigure}
    \begin{subfigure}[t]{.24\linewidth}
    \includegraphics[width=\linewidth]{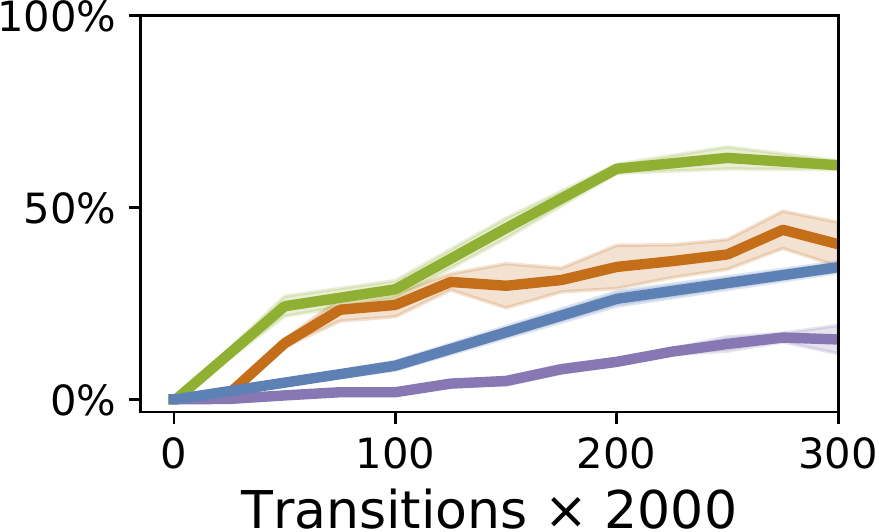}\vspace{-.3em}
    \caption{Throw 4 Objects}
    \end{subfigure}
    \caption{{\bf Downstream task performance in \FPP by \ourMethod{} and the baselines for models checkpointed over the course of free play}.
      We compare \ourMethod{} and \mlpicem, which use the model's epistemic uncertainty approximated via the ensemble disagreement as intrinsic reward, with variants that use a separate RND module. We used three independent seeds.
    }
    \label{fig:task_performance:FPP_wRND}
\end{figure}

\Fig{fig:interactions:FPP_wRND} illustrates that the \gnnrndicem and \mlprndicem both produce less single- and multi-object interactions and flipping behavior than their ensemble disagreement counterparts. In terms of object(s) in air time, \gnnrndicem surpasses \ourMethod. This is expected as once the GNN ensemble learns the lifting behavior, this knowledge is shared among all objects. The GNN ensemble focuses more on bringing two objects together, flipping and/or rolling them. In the case of RND, covering the whole air space with the different cubes is still incentivized since there is no connection between the dynamics model and the RND module. However, this behavior does not necessarily lead to superior task performance for the RND variant, as shown in \fig{fig:task_performance:FPP_wRND}. There is a large difference in the achieved success rates after 300 training iterations, where each training iteration corresponds to collecting 2000 Transitions and training the models. In the case of flipping, \ourMethod is again superior to the \gnnrndicem variant. On the pick \& place task, similar end performance is reached and yet \ourMethod reaches better performance faster. Despite the fact that \gnnrndicem collects a lot of data with objects in air between training iterations 100-200, this doesn't culminate in any significantly better task performance in the pick \& place and stacking tasks, where lifting is a key component. Similarly for the throwing task, we observe the superior performance of the ensemble disagreement-based methods, \ourMethod and \mlpicem, over their RND counterparts.

These results showcase the sample-efficiency of our method not just in terms of generated interactions, but also in terms of zero-shot downstream task performance.

Another important observation in these experiments is the sample-efficiency we obtain through model-based planning alone. If we compare \mlprndicem with the standard RND baseline performance that is trained with an exploration policy as shown in \fig{fig:interactions:FPP}, we get much more interaction-rich exploration during free-play. This again highlights the importance of planning for multi-step intrinsic rewards into the future.

\section{Preliminary Results for \ourMethod in \Robodesk}
\begin{figure}
    \centering
    \includegraphics[width=0.6\linewidth]{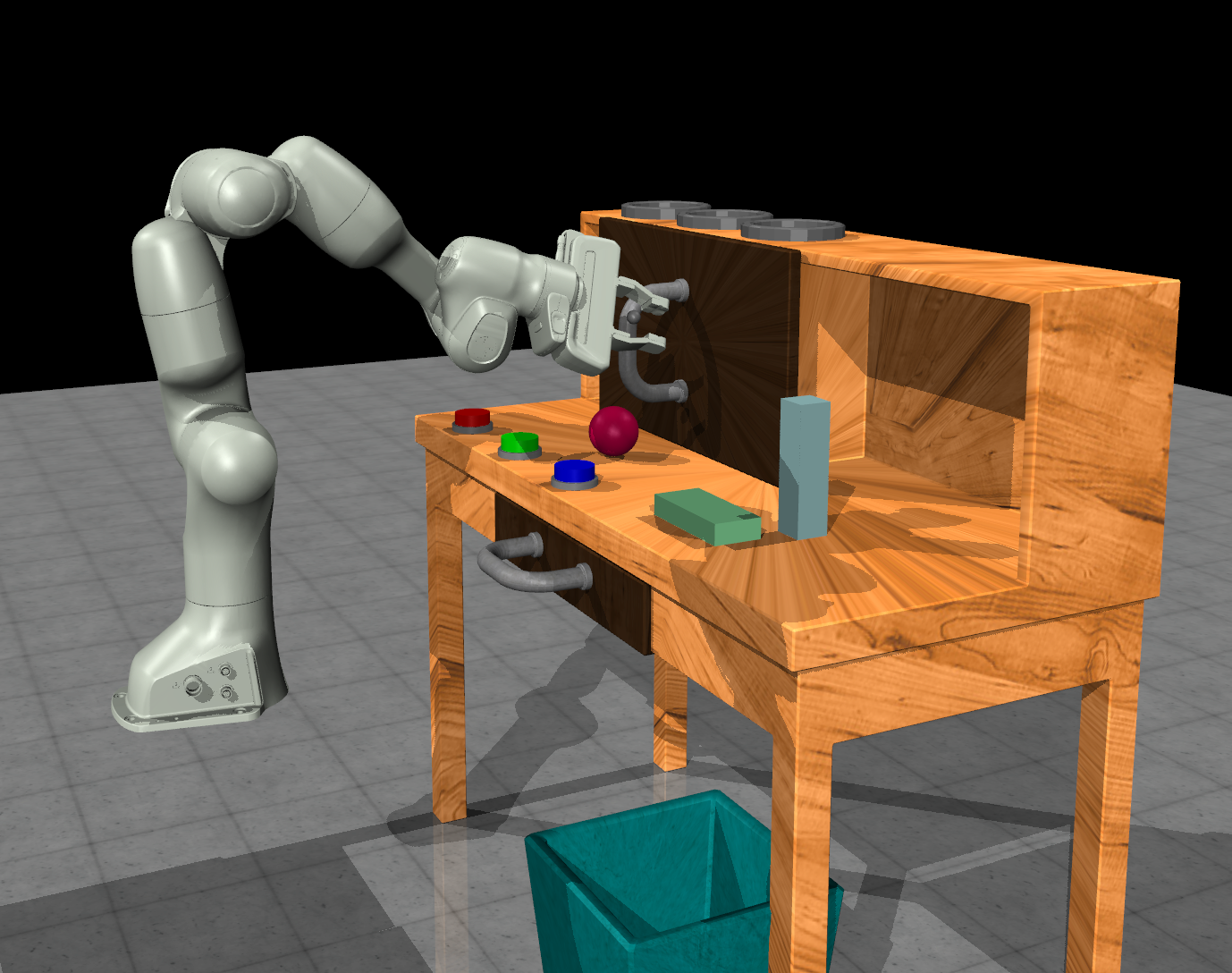}
    \caption{\Robodesk environment.}
    \label{fig:robodesk}
\end{figure}

We apply \ourMethod to the \Robodesk environment \cite{kannan2021robodesk} (\fig{fig:robodesk}) in order to test if our method can deal with diverse geometries of objects with only proprioceptive state information. This environment has complex objects/entities such as a drawer, a sliding cabinet, buttons and other blocks. 

 \begin{table}[!b]
    \centering
  \caption{Preliminary success rates for zero-shot generalization in the extrinsic phase of \ourMethod in \Robodesk.}
    \label{tab:robodesk}
    \renewcommand{\arraystretch}{1.03}
    \resizebox{.9\linewidth}{!}{ %
    \begin{tabular}{@{}l|cccc@{}}
    \toprule
                      & \multicolumn{4}{c}{\textbf{Task}} \\
                      & Open Drawer & Open Slide Cabinet	  & Push Green Button & Push Flat Block Off Table \\
    \midrule
    \ourMethod &  $0.97 \pm 0.02$  &  $0.87 \pm 0.07$  & $0.87 \pm 0.07$  & $0.58 \pm 0.09$ \\
  \bottomrule
    \end{tabular}
   }
\end{table}
For \Robodesk, we encode each entity’s state purely as proprioceptive information of position, quaternion, linear and angular velocities. Note that the entities have even different joint types, where the drawer has a slide/prismatic joint along y-axis, the sliding cabinet a slide joint only in x-axis, the buttons slide joints in z-axis, and the blocks and the ball corresponding to free joints. The different entity types are encoded as static object features and are categorical variables with one-hot encoding.

In our experiments, during the intrinsic phase of \ourMethod{}, the robot arm interacts with the different entities, \eg opening drawer and cabinet, pushing blocks and pushing buttons. The learned GNN ensemble can then be used in the extrinsic phase to solve downstream tasks zero-shot. We test opening the drawer, sliding the cabinet, pushing buttons, and moving blocks yielding the following success rates shown in \tab{tab:robodesk}.

The corresponding videos can be found on our supplementary website \url{https://martius-lab.github.io/cee-us}.

\end{document}